\documentclass[sigconf, nonacm]{acmart}

\usepackage{subcaption}

\newcommand\vldbdoi{XX.XX/XXX.XX}
\newcommand\vldbpages{XXX-XXX}
\newcommand\vldbvolume{14}
\newcommand\vldbissue{1}
\newcommand\vldbyear{2020}
\newcommand\vldbauthors{\authors}
\newcommand\vldbtitle{\shorttitle} 
\newcommand\vldbavailabilityurl{URL_TO_YOUR_ARTIFACTS}
\newcommand\vldbpagestyle{plain} 

\newcommand{\model}{\textsf{\small PILOT-C}}

\begin{document}
\title{PILOT-C: Physics-Informed Low-Distortion Optimal Trajectory Compression}

\author{Kefei Wu}
\affiliation{%
  \institution{Fudan University}
  \country{China}
}
\email{wukf23@m.fudan.edu.cn}

\author{Baihua Zheng}
\affiliation{%
\institution{Singapore Management University}
\country{Singapore}
}
\email{bhzheng@smu.edu.sg}
\author{Weiwei Sun}
\affiliation{%
\institution{Fudan University}
\country{China}
}
\email{wwsun@fudan.edu.cn}

\begin{abstract}

Location-aware devices continuously generate massive volumes of trajectory data, creating demand for efficient compression. Line simplification is a common solution but typically assumes 2D trajectories and ignores time synchronization and motion continuity. 
We propose \model, a novel trajectory compression framework that integrates frequency-domain physics modeling with error-bounded optimization. 
Unlike existing line simplification methods, \model\ supports trajectories in arbitrary dimensions, including 3D, by compressing each spatial axis independently.
Evaluated on four real-world datasets, \model\ achieves superior performance across multiple dimensions. In terms of compression ratio, \model\ outperforms CISED-W, the current state-of-the-art SED-based line simplification algorithm, by an average of 19.2\%.
For trajectory fidelity, \model\ achieves an average of 32.6\% reduction in error compared to CISED-W. Additionally, \model\ seamlessly extends to three-dimensional trajectories while maintaining the same computational complexity, achieving a 49\% improvement in compression ratios over SQUISH-E, the most efficient line simplification algorithm on 3D datasets. 

\end{abstract}

\maketitle

\pagestyle{\vldbpagestyle}
\begingroup\small\noindent\raggedright\textbf{PVLDB Reference Format:}\\
\vldbauthors. \vldbtitle. PVLDB, \vldbvolume(\vldbissue): \vldbpages, \vldbyear.\\
\href{https://doi.org/\vldbdoi}{doi:\vldbdoi}
\endgroup
\begingroup
\renewcommand\thefootnote{}\footnote{\noindent
This work is licensed under the Creative Commons BY-NC-ND 4.0 International License. Visit \url{https://creativecommons.org/licenses/by-nc-nd/4.0/} to view a copy of this license. For any use beyond those covered by this license, obtain permission by emailing \href{mailto:info@vldb.org}{info@vldb.org}. Copyright is held by the owner/author(s). Publication rights licensed to the VLDB Endowment. \\
\raggedright Proceedings of the VLDB Endowment, Vol. \vldbvolume, No. \vldbissue\ %
ISSN 2150-8097. \\
\href{https://doi.org/\vldbdoi}{doi:\vldbdoi} \\
}\addtocounter{footnote}{-1}\endgroup

\ifdefempty{\vldbavailabilityurl}{}{
\vspace{.3cm}
\begingroup\small\noindent\raggedright\textbf{PVLDB Artifact Availability:}\\
The source code, data, and/or other artifacts have been made available at \url{\vldbavailabilityurl}.
\endgroup
}

\section{Introduction}

With the rapid proliferation of location-aware devices (e.g., autonomous vehicles, drones, and IoT sensors), massive volumes of trajectory data are being generated daily. These data are typically collected at specific sampling rates and are transmitted to servers for use in a variety of applications. However, the sheer size of raw trajectory data, often riddled with redundant high-frequency noise and physically implausible motion patterns, presents significant challenges for both storage and real-time processing~\cite{lin2019one,nibali2015trajic,muckell2014compression,douglas1973algorithms,pavlidis1974segmentation}.

Early \emph{lossless} trajectory compression methods, such as delta encoding~\cite{nibali2015trajic,cudre2010trajstore}, achieved only modest compression ratios due to the inherent irregularity of trajectory data at the storage level. While suitable for archival purposes, they fall short in meeting the performance requirements of real-time systems. Consequently, attention has shifted toward \emph{lossy} compression techniques, which strategically trade off some accuracy for higher compression efficiency.

Lossy trajectory compression fall into two main categories: \emph{semantic-driven} and \emph{geometric-driven} methods. 
Semantic-driven methods~\cite{DBLP:journals/pvldb/SongSZZ14,gotsman2015dilution,han2017compress,zhao2018rest,sandu2015spatio,DBLP:journals/tkde/YangWYLZ18} leverage external domain knowledge, such as road networks, to enhance compression ratios. However, their performance is heavily dependent on the availability and accuracy of such contextual information, limiting their generalizability and robustness across domains. 

In contrast, geometric-driven methods are knowledge-agnostic, simpler to implement, and typically provide error-bounded approximations~\cite{muckell2014compression,douglas1973algorithms,lin2017one,chen2009trajectory,chen2012fast,liu2015bounded,liu2016novel,meratnia2004spatiotemporal}.
Among them, line simplification algorithms~\cite{lin2021error} are the most widely studied. These techniques retain only the key points in a trajectory, approximating the path using simplified polylines. Representative techniques, such as CISED~\cite{lin2019one}, achieve strong compression performance under a specified error bound, typically measured using the Synchronized Euclidean Distance (SED). Despite their popularity, line simplification approaches exhibit several fundamental limitations:
%
1) Trajectories are typically treated as geometric curves, abstracted from time. As a result, temporal dynamics (e.g., velocity changes, acceleration) are either ignored or represented only at retained points.
2) High-frequency noise (e.g., GPS jitter) is often retained, while meaningful low-frequency motion patterns (e.g., smooth acceleration or uniform motion) are not effectively preserved.
3) Although constrained by maximum error, their average error is often high, resulting in a distorted representation of the overall trajectory.
4) Most existing methods do not optimize storage structures, leading to inefficiencies in real-world deployment despite good algorithmic performance.

To overcome these challenges, we propose a novel frequency-domain approach to trajectory compression that fundamentally differs from traditional line simplification. Rather than approximating a trajectory with fewer points, we model the trajectory as a multi-dimensional time-series signal and transform it into the frequency domain using the Discrete Cosine Transform (DCT), a technique widely used in image, audio, and video compression due to its ability to separate meaningful low-frequency signals from noise-dominated high-frequency components~\cite{wallace1992jpeg,brandenburg1999mp3,le1991mpeg}. 

\begin{figure}[t]
    \centering
    \begin{subfigure}{0.49 \linewidth}
        \includegraphics[width=\linewidth]{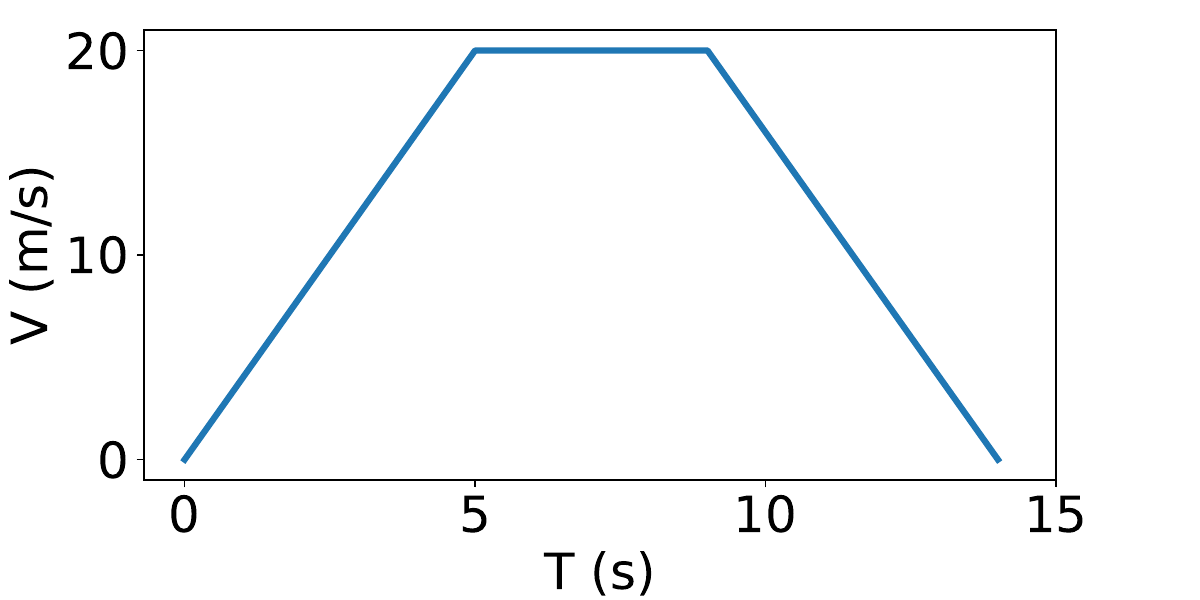}
        \vspace{-0.22in}
        \caption{Velocity-Time curve}
        \label{fig:velocity_time_curve}
    \end{subfigure}
    \hfill
    \begin{subfigure}{0.49 \linewidth}
        \includegraphics[width=\linewidth]{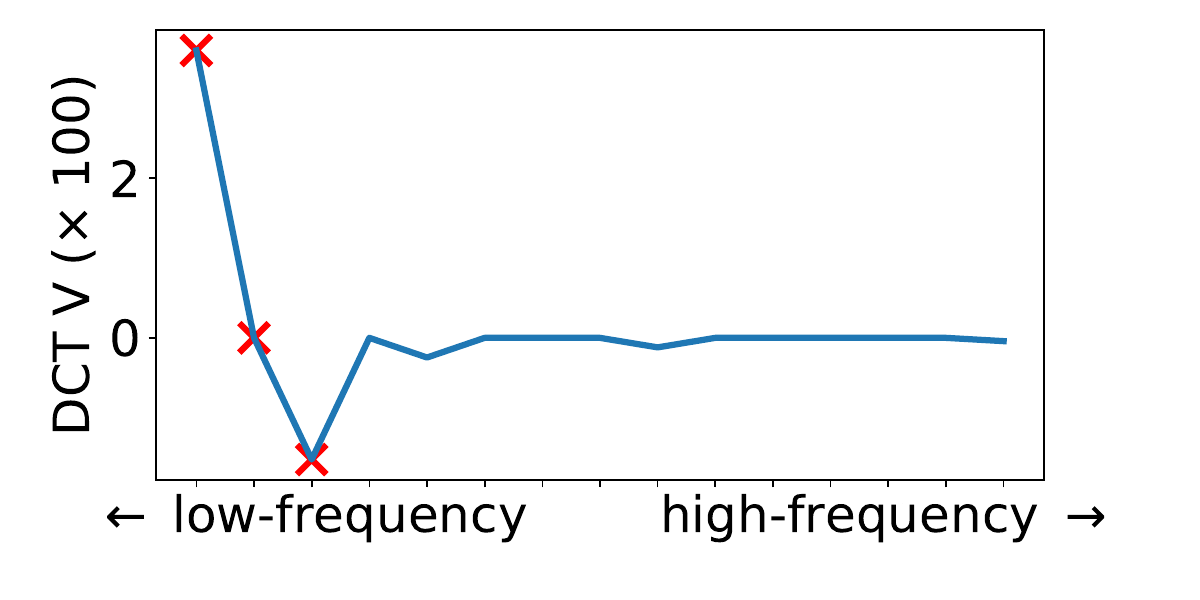}
        \vspace{-0.22in}
        \caption{DCT result}
        \label{fig:DCT_result}
    \end{subfigure}
    \vspace{-0.15in}
    \caption{The velocity-time curve of an object moving in common motion, and the DCT result of the velocity array. 
    }
    \vspace{-0.2in}
\end{figure}

This transformation reveals a key insight: trajectory data, especially from physical systems, exhibit strong temporal and spatial continuity. Objects tend to move smoothly, with gradual changes in velocity and direction, rather than abrupt or oscillatory movements.
As such, the high-frequency components of the DCT-transformed trajectory often represent noise or minor irregularities. These components can be selectively discarded with minimal impact on the perceived shape or dynamics of the motion, enabling high compression ratios with low distortion.
Figure~\ref{fig:velocity_time_curve} shows a common velocity-time graph in which an object accelerates from rest, maintains a constant speed, and then decelerates to a stop. After applying DCT, as shown in Figure~\ref{fig:DCT_result}, the velocity signal is decomposed into a few dominant low-frequency components labeled by cross and many minor high-frequency ones. 

To operationalize this insight, we propose \model\ (\textit{\underline{P}hysics-\underline{I}nformed \underline{L}ow-Distortion \underline{O}ptimal \underline{T}rajectory \underline{C}ompression}), a new framework that integrates physics modeling of motion with frequency-domain signal processing and error-bounded optimization. Our contributions are as follows: 
\begin{itemize}
\item{\bf Frequency-domain, physics-informed compression.} We are the first to exploit the physical structure of motion for trajectory compression. 
By applying DCT to each spatial dimension independently, we capture and retain the dominant low-frequency components that encode key motion behaviors. This enables us to compress trajectories effectively by removing high-frequency components that correspond to noise or physically implausible transitions.

\item{\bf Scalability to higher dimensions.} Unlike many existing algorithms—such as CISED—that are inherently restricted to 2D spatial data, \model\ operates independently on each dimension. This design enables natural extension to higher-dimensional trajectories (e.g., 3D), such as those from aerial drones or motion-capture systems, without redesigning the core algorithm.

\item{\bf Optimized storage and practical integration.} We optimize the storage structure used for storing the compressed representation, making \model\ both theoretically sound and practically deployable. This storage structure is general and can be integrated into existing line simplification algorithms to enhance their compression ratios in real-world applications.  

\item{\bf Comprehensive experimental evaluation.} We evaluate \model\ on four real-world datasets (nuPlan, GeoLife, Mopsi, GeoLife-3D) against leading methods (CISED-S, CISED-W, SQUISH). Under the same SED constraint, it improves compression rate by 19.2\% and reduces average SED error by 32.6\%, preserving trajectory structure more effectively.
\end{itemize}

The rest of this paper is organized as follows.  Section~\ref{sec:preliminaries} introduces foundational concepts and background. Section~\ref{sec:model} presents our trajectory compression framework \model. Section~\ref{sec:experimental_study} reports the experimental results and analysis.
Section~\ref{sec:related_work} reviews related work, and Section~\ref{sec:conclusions} concludes this paper.

\section{Preliminaries}
\label{sec:preliminaries}

In this section, we introduce the core of our algorithm: the Discrete Cosine Transform (DCT). We then present a point encoding technique, which is applied in both the traditional line simplification algorithms and our proposed \model\ to reduce storage.

\subsection{Discrete Cosine Transform (DCT)}
\label{subsec:DCT}

The Discrete Cosine Transform (DCT)~\cite{DBLP:journals/tc/AhmedNR74} is a real-valued, orthogonal transformation widely used in signal processing for data compression. Among its variants, the Type-II DCT (DCT-II) is the most common. For a one-dimensional discrete signal $X_n$, where $n=0,1,\ldots,N-1$, the DCT-II is defined as:
\begin{equation}
    \resizebox{0.85\linewidth}{!}{
    $
    C_k=\alpha_k \sum\nolimits_{n=0}^{N-1}X_n\cos\left(\frac{\left(2n+1\right)k\pi}{2N}\right),
    k=0,1,\ldots,N-1
    $}
\end{equation}
where normalization factor $\alpha_0=\sqrt{\frac{1}{N}}$ and $\alpha_k=\sqrt{\frac{2}{N}}$ for $k>1$. 
%

The Inverse Discrete Cosine Transform (IDCT) reconstructs the original sequence $X_n$ :
\begin{equation}
    \resizebox{0.85\linewidth}{!}{
    $
    X_n=\sum\nolimits_{k=0}^{N-1}\alpha_k C_k\cos\left(\frac{\left(2n+1\right)k\pi}{2N}\right),n=0,1,\ldots,N-1
    $}
\end{equation}
%

To better suit our application, in our implementation, 1) for DCT, we set $\alpha_k = 2$ for $k \geq 1$; and 2) for IDCT, we use $\alpha_k = \frac{1}{N}$. We leave $\alpha_0$ unmodified, since $C_0$ is always zero in our algorithm. These modifications remove the need for square root and division operations in IDCT. 

Though DCT-II is theoretically  lossless when paired with its inverse and computed with full precision, practical applications involve quantization - especially of high-frequency coefficients. This introduces loss and enables compression.
%
A core advantage of the DCT is its  energy compaction: it concentrates most of a signal's energy into a few low-frequency coefficients, making it well-suited for compressing highly correlated data by retaining significant components and discarding negligible ones.

To reduce computational complexity from $O(n^2)$ to $O(n\log n)$, we use fast DCT algorithms \cite{loeffler1989practical, arai1988fast} that map the DCT to a Fast Fourier Transform (FFT). This involves symmetrically extending an $N$-point signal to $2N$ points, applying FFT, and extracting the scaled real part to obtain the DCT coefficients. 

\subsection{Point quantization and encoding}
\label{subsec:point-encoding}
While trajectory simplification has been well-studied, the efficient storage of simplified points remains underexplored, often leading to suboptimal compression in practice. To address this, we propose a point quantization and encoding scheme inspired by delta encoding for more effective storage of compressed signals. This general-purpose scheme can be seamlessly integrated with existing line simplification algorithms, reducing their storage overhead and enabling fair, storage-aware comparisons with our approach. 

\begin{figure}[t]
    \centering
    \includegraphics[width=\linewidth]{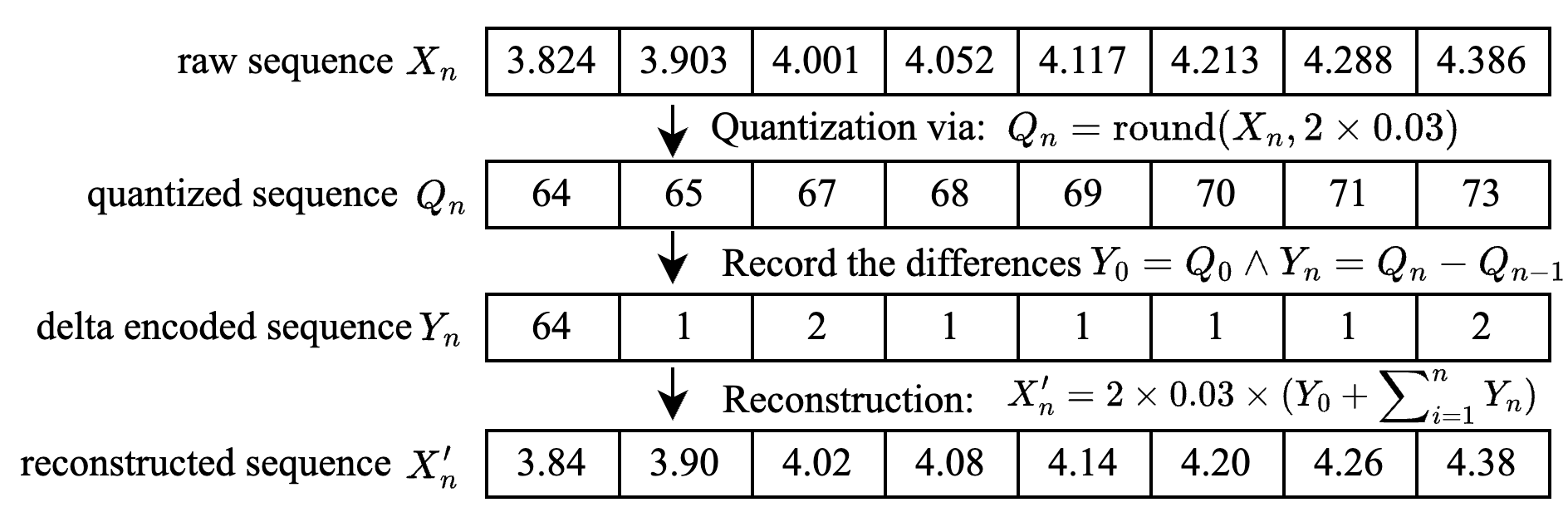}
    \vspace{-0.3in}
    \caption{Example of Quantization ($\epsilon_d=0.03$)
    }
    \label{fig:quantization}
    \vspace{-0.15in}
\end{figure}

Given a floating-point signal sequence $X_n$ (see Figure~\ref{fig:quantization}), we first apply quantization using a predefined error bound $\epsilon_d$. A rounding function, $\text{round}(X_n, 2\epsilon_d)$, maps each value $X_n$ to the nearest multiple of $2\epsilon_d$, producing an integer $X'_n$. The original value can be re-constructed approximately via $2\epsilon_d \times X'_n$, with a construction error bounded by $\epsilon_d$, i.e., $|X_n-X'_n|\le \epsilon_d$. 
For example, with $\epsilon_d=0.03$, the value $x=3.824$ becomes 64, while $64\times (2\times \epsilon_d)=3.84$, within $\pm 0.03$ of the original. 
The resulting integer sequence $Q_n$ is delta-encoded as $Y_0=Q_0$ and $Y_n=Q_n-Q_{n-1}$ for $n\ge 1$, shown in Figure~\ref{fig:quantization}. If the raw sequence $X_n$ contains only integers, quantization can be skipped, and we simply store the first value and differences from the previous values for subsequent values for a more compact representation. 

%
%

\begin{figure}[t]
\vspace{-0.1in}
    \centering
    \includegraphics[width=\linewidth]{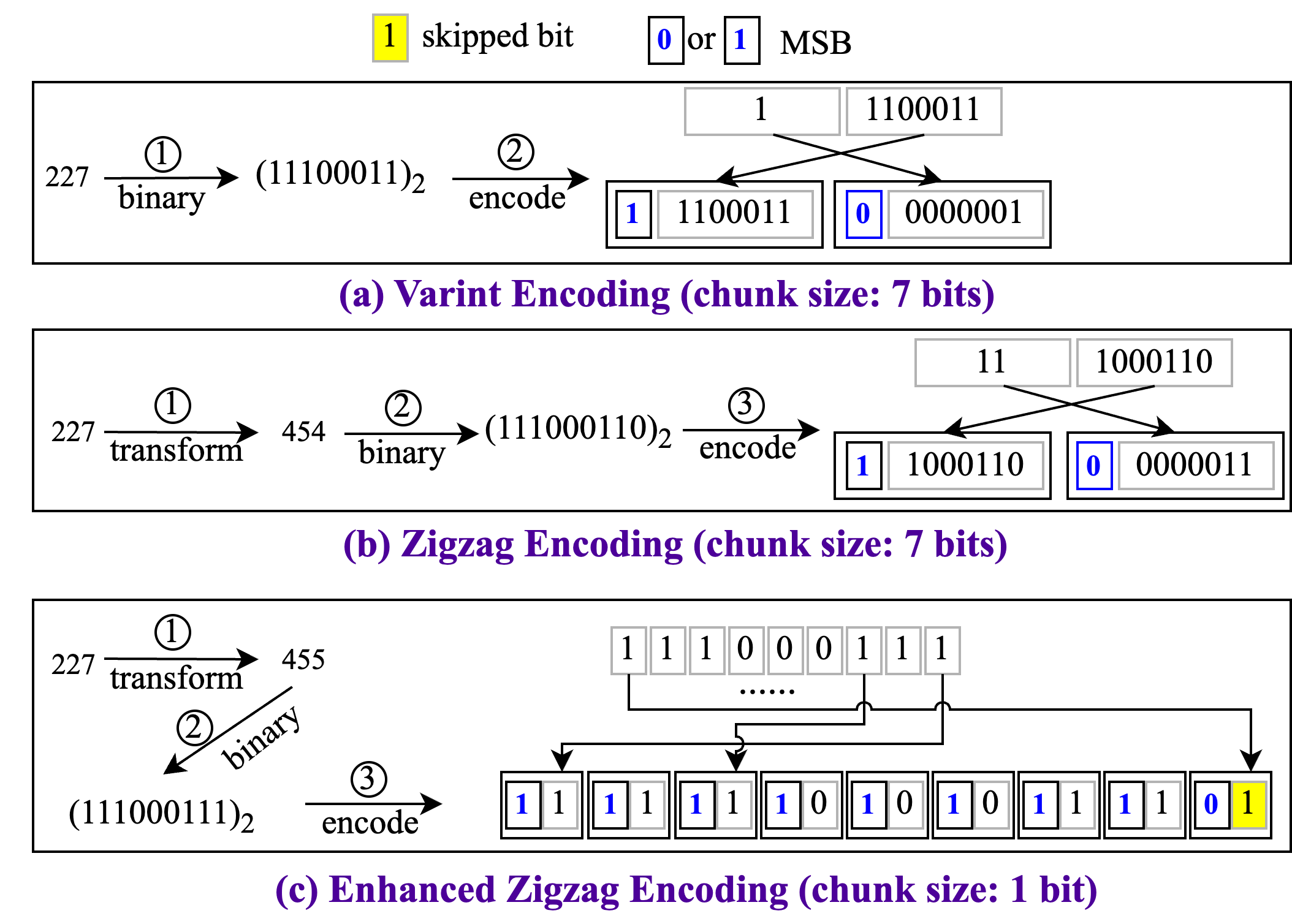}
    \vspace{-0.25in}
    \caption{Examples and explanations of Varint, Zigzag, and our Enhanced Zigzag for input number 227. 
    }
    \label{fig:encoding-v2}
    \vspace{-0.2in}
\end{figure}

To further compress $Y_n$, we apply Varint encoding, a variable-length binary encoding scheme optimized for integers, and its variant Zigzag encoding for signed numbers. Varint splits each integer into 7-bit chunks, with each chunk stored in a byte. The most significant bit (MSB) of each byte acts as a continuation flag: a MSB $=1$ indicates that more bytes follow, and MSB $=0$ marks the final byte in the sequence. For example, the integer $227$ (binary 11100011) is split into two chunks: 1 and 1100011. Varint encodes these in reverse order:  \textbf{1}\underline{110 0011} (MSB $=1$, value $=99$) and \textbf{0}\underline{000 0001} (MSB $=0$, value $=128$), as shown in Figure~\ref{fig:encoding-v2}(a). Decoding sums these values to recover $227$ ($=99+128$).


To efficiently encode signed integers, we use Zigzag encoding, which remaps signed integers to unsigned ones. It transforms non-negative numbers to even numbers ($n \xrightarrow{} 2n$) and negative ones to odd numbers ($n \xrightarrow{} 2|n| - 1$). This mapping converts small-magnitude numbers - whether positive or negative - to small unsigned integers, which Varint encodes compactly. In our example shown in Figure~\ref{fig:encoding-v2}(b), Zigzag transforms input number 227 to $2\times 227=454$ and then encodes 454 with Varint.

While traditional Varint + Zigzag encoding works well in general, it has two key limitations. First, traditional implementations fix the chunk size (e.g., seven bits in our examples), which may not suit all datasets. We address this by introducing a tunable chunk length parameter $l$, configurable at the algorithm's initialization to better adapt to different datasets. 
%
Second, standard implementations introduce redundancy: since the final byte in the encoding sequence must have MSB $=0$, meaning sequences ending in an all-zero byte are wasted. This redundancy becomes more prominent with small chunk size (e.g., 1 bit), where all encoded sequences will end in '01' where 0 is MSB, and 1 is the real value. 

To reduce this redundancy, we propose an enhanced Zigzag encoding variant for specific use cases. This version shifts the mapped values to eliminate zeros entirely: non-negative numbers are mapped as $n \xrightarrow{} 2n+1$ and negative ones as ($n \xrightarrow{} 2|n|$). This ensures that all remapped integers begin with a binary bit `1'. When $l=1$, the final encoded sequence always ends in `1', which can be safely omitted during storage and restored during decoding. 
For example, the input 227 becomes $2\times 227+1=455$ (binary 111000111). If we set the chunk size to 1 bit, Varint encoding produces a sequence where the final bit `1' can be skipped, as shown in Figure~\ref{fig:encoding-v2}(c), reducing total storage.  

\section{Model}
\label{sec:model}

This section introduces the \model\ compression framework, analyzing its correctness and complexity. We then propose a truncation procedure to further reduce storage and a post-compression validation step to correct any misalignments between the original and compressed points that exceed the error bound. Next, we describe the proposed storage structure for storing compressed trajectories and the decompression procedure. Finally, we extend \model\ to support trajectories with arbitrary sampling frequencies and discuss its strengths and limitations.
Table~\ref{tab:symbols} summarizes the frequently used notations.

\begin{table}\label{tab:symbol}
	\centering
\renewcommand\arraystretch{0.95}
	\caption{{Symbols and description}}\vspace{-4mm}
	\label{tab:symbols}
	\small
	\setlength{\tabcolsep}{3pt}
	\begin{tabular}{p{1cm}p{7cm}}
		\hline
		\textbf{Notation} & \textbf{Description} \\
        $\mathcal{T}$     & a raw trajectory that contains $|\mathcal{T}|$ elements, where each element $(p_i, t_i)$ captures the position $p_i$ at timestamp $t_i$ \\ 
        $\mathcal{S}$, $\mathcal{S}_{i,j}$ & a signal array, the $j$-th signal array in $i$-th spatial dimension \\ 
        $\mathcal{V}$    & a velocity array derived based on an input signal array \\ 
        $\mathcal{C}$     & a frequency array produced by transforming a velocity array via DCT \\
        $\mathcal{F}$  & output frequency array via rounding and encoding $\mathcal{C}$\\
        $\mathcal{C}'$,$\mathcal{V}'$,$\mathcal{S}'$ & arrays recovered by reconstructing a trajectory through decompressing the output frequency array $\mathcal{F}$ generated by \model \\
        $b_s$   &   block size to segment signal sequences \\
        $dim$   &   total number of spatial dimensions in a trajectory\\
        $\epsilon$ & the overall error bound of a compression algorithm\\
        $\epsilon_p$ & maximum allowable point-wise error \\
        $\epsilon_f$ &  maximum allowable frequency error \\
        $\epsilon_t$ &  required temporal precision \\
        \hline
	\end{tabular}
	\vspace{-0.2in}
\end{table}
\setlength{\textfloatsep}{15pt}

\subsection{Compression framework}
\label{subsec:framework}


\begin{figure*}[h]
    \centering
    \includegraphics[width=\linewidth]{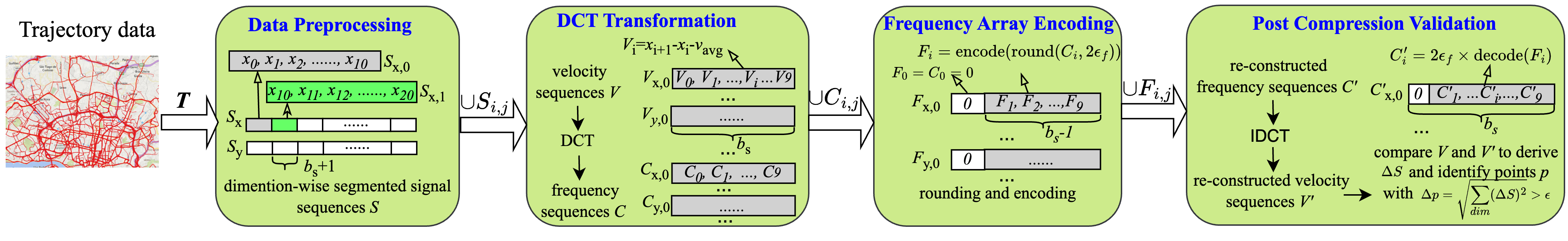}
    \vspace{-0.3in}
    \caption{ Physics-Informed Low-Distortion Optimal Trajectory Compression framework ($b_s=10$).
    }
    \label{fig:model_structure2}
    \vspace{-0.2in}
\end{figure*}



Our compression model \model\ processes raw trajectory data to generate a compact representation. Without loss of generality, we represent a trajectory $\mathcal{T}$ as $\mathcal{T}=\langle (p_0, t_0), (p_1, t_1), \cdots\rangle$, where each tuple $(p_i, t_i)$ captures the position $p_i$ of the moving object at timestamp $t_i$. The structure of each point $p_i$ is flexible. For example, for vehicles on road networks, $p_i$ consists of longitude $x_i$ and latitude $y_i$; for drones or aerial vehicles, $p_i$ includes longitude $x_i$, latitude $y_i$, and altitude $z_i$. Alternatively, positions can be represented in the spatial geodetic Cartesian coordinate system using $x_i$, $y_i$, $z_i$.

To simplify the discussion, we first assume uniform sampling frequency, meaning the time interval $\Delta t$ between consecutive points in a trajectory is constant: $\forall (p_i,t_i)\in \mathcal{T}$, $t_i-t_{i-1}=t_{i+1}-t_i=\Delta t$. 
We will later relax this assumption and explain how the model can accommodate trajectories with non-uniform sampling intervals. 

Our model consists of three key stages: \emph{data pre-processing}, \emph{DCT transformation}, and \emph{frequency array encoding}, as detailed below. Figure~\ref{fig:model_structure2} provides a visual overview of the compression process.
\noindent
\textbf{Data preprocessing.}
We first decompose each input trajectory into signal sequences $\mathcal{S}$ along each spatial dimension. For example, given a trajectory $\mathcal{T}$, the signal sequence $\mathcal{S}$ along the $x$-dimension is: $\langle(x_0, t_0), (x_1, t_1), \cdots \rangle$, and similarly for $y$- and $z$-dimensions if applicable. Since time intervals are uniformly spaced, we omit them and represent the signal sequence $\mathcal{S}$ as a one-dimensional array: $\mathcal{S}=(S_0, S_1, \cdots)$, where, for example, $S_0=x_0$, $S_1=x_1$ and so on for the signal sequence along the $x$-dimension.
Each signal sequence corresponding to the $i$-th spatial dimension is then partitioned into blocks $\mathcal{S}_{i,j}$ of based on block size $b_s$, which is a tunable parameter.  
To maintain continuity between blocks, the last point of each signal block is reused as the starting point of the next. For instance, $\mathcal{S}_{x,0}=(x_0, x_1,\cdots, x_{b_s})$, $\mathcal{S}_{x,1}=(x_{b_s},x_{b_s+1}, \cdots, x_{2b_s})$, and so on. 
This segmentation improves compression efficiency, as demonstrated in our experiments. 
When the context is clear, we use $\mathcal{S}$ to represent a signal array of $b_s+1$ size corresponding to a single spatial dimension.

\noindent
\textbf{DCT transformation}.
%
%
The difference between two consecutive signal values approximates the velocity of the object. Since velocity changes more smoothly than position, 
it is more suitable for DCT processing. 
Within each block and for each spatial dimension, we compute the velocity $v_i$ as the difference between consecutive signal values: $v_i=S_{i+1}-S_{i}$. 
%
For example, along the $x$-dimension, $v_i=x_{i+1}-x_i$. We also calculate the average velocity $v_{avg}$ within the block: $v_{avg}=\frac{\sum\nolimits_{i=0}^{b_s-1}v_i}{b_s}=\frac{S_{b_s}-S_0}{b_s}$. 
Similar to delta compression, we construct a zero-centered velocity array $\mathcal{V}=(V_0, V_1, \cdots, V_{b_s-1})$ by subtracting the average velocity from each velocity value, where $V_0=v_0-v_{avg}$, $V_1= v_1-v_{avg}$, etc. By construction, the sum of all values in $\mathcal{V}$ is zero: $\sum V_i=0$. Note the length of each velocity array $\mathcal{V}$ is $b_s$. 
Each velocity array $\mathcal{V}$ is then transformed using DCT, producing a frequency array $\mathcal{C}=(C_0, C_1, \cdots, C_{b_s-1})$. 
Since the input $\mathcal{V}$ is zero-centered, the DC component $C_0$ in $\mathcal{C}$ that represents the average of the input signal is zero: $C_0=0$. 

\noindent
\textbf{Frequency array encoding}. 
To compress the frequency array $\mathcal{C}=(C_0, C_1, \cdots, C_{b_s-1})$, we quantize each frequency component based on predefined frequency error threshold $\epsilon_f$ that defines the maximum allowed frequency error. Each component $C_i\in \mathcal{C}$ is quantized as $\text{round}({C_i},{2\epsilon_f})$, 
where $\text{round}$ is the rounding function that maps each value $C_i$ to the nearest multiple of $2\epsilon_f$.
Each quantized component is then encoded using the enhanced Zigzag encoding. 
Let $\mathcal{F}$ denote the rounded frequency array, $\mathcal{F}=(F_1, \cdots, F_{b_s-1})$ where $F_i=\text{encode}(\text{round}({C_i}, {2\epsilon_f}))$. Note $F_0=C_0=0$ is not stored in $\mathcal{F}$ to save space.

\noindent
\textbf{Summary}. Given a raw trajectory $\mathcal{T}$ sampled at a uniform rate, our proposed framework, \model, outputs a set of encoded frequency arrays  $\cup_{i<dim, j<\lceil\frac{\mathcal{|T|}-1}{b_s}\rceil}\mathcal{F}_{i,j}$, where each $\mathcal{F}_{i,j}$ represents the encoded frequency coefficients of the $j$-th block along the $i$-th spatial dimension. Our model also stores minimal metadata necessary for accurate decompression, as detailed in Section~\ref{subset:storage}. \model\ naturally supports trajectories in arbitrary dimensions by treating each spatial dimension independently in both compression and reconstruction. This makes our framework particularly suitable for modern applications involving 3D and even higher-dimensional movement data.

\subsection{Correctness and complexity analyses}
\label{subsec:correctness}
In the following, we prove that all reconstruction errors remain within the predefined error bound $\epsilon$, and that the recovered points, reconstructed from the output arrays $\mathcal{F}$ of the compression model and their metadata, faithfully represent the original raw trajectory within that bound $\epsilon$.

Consider a signal array $\mathcal{S}$ representing a trajectory along one spatial dimension within one single block, along with its corresponding velocity array $\mathcal{V}$ and frequency array $\mathcal{C}$. Let $\mathcal{F}$ denote the output array produced by our compression model \model. Given $\mathcal{F}$, we reconstruct the frequency, velocity, and signal arrays, denoted as $\mathcal{C}'$, $\mathcal{V}'$, and $\mathcal{S}'$, respectively. 

\vspace{0.02in}
\noindent
\textbf{Analysis of $\Delta\mathcal{C}$.}
We first analyze the errors caused by rounding up the frequency array $\mathcal{C}$ as follows. 
Given $\mathcal{F}$ and the specific rounding function, each component $C'_{k}$ in $C'$ can be derived via Equation~\eqref{eq:c'_k}. 
\begin{equation}
\resizebox{0.91\linewidth}{!}{
    $
    \begin{aligned}
    C'_{k} = 2\epsilon_f \text{decode}\left(F_k\right) = 2\epsilon_f \text{round}\left({C_k}, {2\epsilon_f}\right), k=1,2,\ldots,b_s-1
    \end{aligned}
    $}
    \label{eq:c'_k}
\end{equation}
Therefore, the element-wise difference between $\mathcal{C}'$ and original array $\mathcal{C}$ is bounded by predefined allowable frequency error $\epsilon_f$:
\begin{equation}
    \Delta C_{k} = C_{k} - C'_{k} \in [-\epsilon_f, \epsilon_f), k=1,2,\ldots,b_s-1
\end{equation}

\noindent
\textbf{Analysis of $\Delta \mathcal{V}$.}
Let $\mathcal{V}'$ denote the recovered velocity array obtained from $\mathcal{C}'$ via IDCT. We calculate the element-wise error in the velocity array $\mathcal{V}$, as listed in Equation~\eqref{eq:delta_v}. In this analysis, we rely on two key properties: i) $\sum V=0$ and $C_0=C_0'=0$; 
and ii) $\alpha'_i = \frac{1}{b_s}$, as stated in Section~\ref{subsec:DCT}.
%
\begin{equation}
\resizebox{0.85\linewidth}{!}{
    $
    \begin{aligned}
        \Delta V_{k} = V_{k} - V'_k 
        &= \sum\nolimits_{i=0}^{b_s-1}\alpha'_i\left(C_i - C'_i\right)\cos\left(\frac{\left(2k+1\right)i\pi}{2b_s}\right) \\
        &= \sum\nolimits_{i=1}^{b_s-1}\frac{\Delta C_i\cos\left(\frac{\left(2k+1\right)i\pi}{2b_s}\right)}{b_s} , k=0,1,\ldots,b_s-1
    \end{aligned}
    $}
    \label{eq:delta_v}
\end{equation}
%

\vspace{0.02in}
\noindent
\textbf{Analysis of $\Delta \mathcal{S}$.}
Next, we derive the element-wise difference between each recovered signal value $S'_k\in \mathcal{S}'$ and its original signal value $S_k\in \mathcal{S}$.
Recall that $S_{k} = S_0+\sum\nolimits_{i=0}^{k-1} V_{i}$, $S'_{k} = S'_0+\sum\nolimits_{i=0}^{k-1} V'_{i}$. Therefore, the difference $\Delta S_{k}=S_{k}-S'_{k}$ can be expressed as $S_0-S'_0+\sum\nolimits_{i=0}^{k-1}\Delta V_i$. Since $S_0-S'_0=0$ (the starting value of each signal array is stored as part of the essential block-level metadata, see Section~\ref{subset:storage}), we simplify $\Delta S_k$ via Equation~\eqref{equ:change_s}.

\begin{equation}
\resizebox{0.80\linewidth}{!}{
    $
    \begin{aligned}
        \Delta S_{k} = \sum\nolimits_{i=0}^{k-1} \Delta V_{i} 
        &= \sum\nolimits_{i=0}^{k-1}\sum\nolimits_{j=1}^{b_s-1}\frac{\Delta C_j\cos\left(\frac{\left(2i+1\right)j\pi}{2b_s}\right)}{b_s} \\
        &= \sum\nolimits_{j=1}^{b_s-1}\frac{\Delta C_j \sum\nolimits_{i=0}^{k-1}\cos\left(\frac{\left(2i+1\right)j\pi}{2b_s}\right)}{b_s} \\
        &= \sum\nolimits_{j=1}^{b_s-1}\frac{\Delta C_j \sin\left(\frac{kj\pi}{b_s}\right)}{2b_s\sin\left(\frac{j\pi}{2b_s}\right)} , k= 1,\ldots,b_s
    \end{aligned}
    $}
    \label{equ:change_s}
\end{equation}

\noindent
\textbf{Proof of Equation~\eqref{equ:change_s}} In this formula, we use 

\begin{equation}
    \sum_{i=0}^{k-1}\cos\left(\frac{\left(2i+1\right)j\pi}{2b_s}\right) = \frac{\sin\left(\frac{kj\pi}{b_s}\right)}{\sin\left(\frac{j\pi}{2b_s}\right)}
\end{equation}

This is because by Product-to-Sum formulas, we have:

\begin{equation}
    \cos\left(a\right)\sin\left(b\right)=\dfrac{1}{2}\left(\sin\left(a+b\right)-\sin\left(a-b\right)\right)
\end{equation}

Let $x=\dfrac{j\pi}{2b_s}$, we can obtain:

\begin{equation}
\resizebox{0.8\linewidth}{!}{
    $
    \begin{aligned}
        2\sum_{i=0}^{k-1}\cos\left(\left(2i+1\right)x\right)\sin\left(x\right) &= \sum_{i=0}^{k-1}\left(\sin\left(\left(2i+2\right)x\right)-\sin\left(2ix\right)\right) \\
        &=\sin\left(2kx\right)
    \end{aligned}
    $}
\end{equation}

We assume that each $\Delta C_{k}$ is independently and uniformly distributed. This implies that $\Delta S_{k}$ is a sum of $k$ independent, uniformly distributed variables. 
From the properties of the uniform distribution, the variance of each $\Delta C_{k}$ is: $\text{Var}\left(\Delta C_j\right) = \frac{\epsilon_f^2}{3}$. Since the expectation of $\Delta S_{k}$ is zero, its variance can be calculated as the sum of the variances of each uniformly distributed variable, as detailed in Equation~\eqref{eq:var_s}.

\begin{equation}
\resizebox{0.9\linewidth}{!}{
    $
    \begin{aligned}
        \text{Var}\left(\Delta S_{k}\right) &= \sum\nolimits_{j=1}^{b_s-1}\text{Var}\left(\frac{\Delta C_j \sin\left(\frac{kj\pi}{b_s}\right)}{2b_s\sin\left(\frac{j\pi}{2b_s}\right)}\right) \\
        &= \sum\nolimits_{j=1}^{b_s-1}\left(\frac{\sin\left(\frac{kj\pi}{b_s}\right)}{2b_s\sin\left(\frac{j\pi}{2b_s}\right)}\right)^2\text{Var}\left(\Delta C_j\right) \\
        &= \sum\nolimits_{j=1}^{b_s-1}\frac{\sin^2\left(\frac{kj\pi}{b_s}\right)}{4b_s^2\sin^2\left(\frac{j\pi}{2b_s}\right)}\frac{\epsilon_f^2}{3} 
    = \frac{\epsilon_f^2}{12b_s^2}\sum\nolimits_{j=1}^{b_s-1}\frac{\sin^2\left(\frac{kj\pi}{b_s}\right)}{\sin^2{\frac{j\pi}{2b_s}}} \\
        &= \frac{\left(kb_s-k^2\right)\epsilon_f^2}{6b_s^2} , k=1,\ldots,b_s
    \end{aligned}
    $}
    \label{eq:var_s}
\end{equation}

\noindent
\textbf{Proof of Equation~\eqref{eq:var_s}} In this formula, we use

\begin{equation}
    \sum_{j=1}^{b_s-1}\frac{\sin^2\left(\frac{kj\pi}{b_s}\right)}{\sin^2{\left(\frac{j\pi}{2b_s}\right)}} = 2kb_s - 2k^2
\end{equation}

First, we prove that $\forall x\in \mathbb{R},m\in\mathbb{N}^{*}$

\begin{equation}
    \sum_{l=0}^{m}\dfrac{\sin\left(\left(l+\frac{1}{2}\right)x\right)}{\sin\left(\frac{1}{2}x\right)}=\dfrac{\sin^2\left(\left(\frac{m+1}{2}\right)x\right)}{\sin^2\left(\frac{1}{2}x\right)}
\end{equation}

This is because

\begin{equation}
\resizebox{0.9\linewidth}{!}{
    $
    \begin{aligned}
        \sum_{l=0}^{m}\dfrac{\sin\left(\left(l+\frac{1}{2}\right)x\right)}{\sin\left(\frac{1}{2}x\right)}&=\sum_{l=0}^{m}\dfrac{\sin\left(\left(l+\frac{1}{2}\right)x\right)\sin\left(\frac{1}{2}x\right)}{\sin^2\left(\frac{1}{2}x\right)} \\
        &=\sum_{l=0}^{m}\dfrac{-\frac{1}{2}\left(\cos\left(\left(l+1\right)x\right)-\cos \left(lx\right)\right)}{\sin^2\left(\frac{1}{2}x\right)} \\
        &=\dfrac{-\frac{1}{2}\left(\cos\left(\left(m+1\right)x\right)-1\right)}{\sin^2\left(\frac{1}{2}x\right)} =\dfrac{\sin^2\left(\left(\frac{m+1}{2}\right)x\right)}{\sin^2\left(\frac{1}{2}x\right)}
    \end{aligned}
    $}
\end{equation}

Let $x=\dfrac{j\pi}{2b_s}$, we can obtain:

\begin{equation}
    \dfrac{\sin^2\left(\frac{kj\pi}{b_s}\right)}{\sin^2\left(\frac{j\pi}{2b_s}\right)}=\sum_{l=0}^{2k-1}\dfrac{\sin{\left(\left(l+\frac{1}{2}\right)\frac{j\pi}{b_s}\right)}}{\sin\left(\frac{j\pi}{2b_s}\right)}
\end{equation}

Next, we prove that $\forall x\in \mathbb{R},m\in\mathbb{N}^{*}$

\begin{equation}\label{1}
    \dfrac{\sin\left(\left(m+\frac{1}{2}\right)x\right)}{\sin\left(\frac{1}{2}x\right)}=1+2\sum_{l=1}^{m}\cos\left(lx\right)
\end{equation}

This is because

\begin{equation}
\resizebox{0.7\linewidth}{!}{
    $
    \begin{aligned}
        & \left(1+2\sum_{l=1}^{m}\cos\left(lx\right)\right)\sin\left(\dfrac{1}{2}x\right) \\
        =&\sin\left(\dfrac{1}{2}x\right)+2\sum_{l=1}^{m}\cos\left(lx\right)\sin\left(\dfrac{1}{2}x\right) \\
        =&\sin\left(\dfrac{1}{2}x\right)+\sum_{l=1}^{m}\left(\sin\left(\left(l+\dfrac{1}{2}\right)x\right)-\sin\left(\left(l-\dfrac{1}{2}\right)x\right)\right) \\
        =&\sin\left(m+\dfrac{1}{2}\right)x
    \end{aligned}
    $}
\end{equation}

So that


\begin{equation}
\resizebox{0.8\linewidth}{!}{
    $
    \begin{aligned}
        \sum_{j=1}^{b_s-1}\dfrac{\sin^2\left(\frac{kj\pi}{b_s}\right)}{\sin^2\left(\frac{j\pi}{2b_s}\right)} &=\sum_{j=1}^{b_s-1}\sum_{l=0}^{2k-1}\dfrac{\sin{\left(\left(l+\frac{1}{2}\right)\frac{j\pi}{b_s}\right)}}{\sin\left(\frac{j\pi}{2b_s}\right)} \\
        &=\sum_{j=1}^{b_s-1}\sum_{l=0}^{2k-1}\left(1+2\sum_{m=1}^{l}\cos\left(\dfrac{mj\pi}{b_s}\right)\right) \\
        &=2\sum_{j=1}^{b_s-1}\sum_{l=0}^{2k-1}\sum_{m=1}^{l}\cos\left(\dfrac{mj\pi}{b_s}\right)+2kb_s-2k
    \end{aligned}
    $}
\end{equation}

Then we need to calculate $\sum_{j=1}^{b_s-1}\sum_{l=0}^{2k-1}\sum_{m=1}^{l}\cos\left(\dfrac{mj\pi}{b_s}\right):$

By exchanging the sum order, we can get:

\begin{equation}
    \sum_{j=1}^{b_s-1}\sum_{l=0}^{2k-1}\sum_{m=1}^{l}\cos\left(\dfrac{mj\pi}{b_s}\right)=\sum_{l=0}^{2k-1}\sum_{m=1}^{l}\sum_{j=1}^{b_s-1}\cos\left(\dfrac{mj\pi}{b_s}\right)
\end{equation}

If $m$ is odd, then 

\begin{equation}
    \cos\left(\dfrac{mj\pi}{b_s}\right)+\cos\left(\dfrac{m\left(b_s-j\right)\pi}{b_s}\right)=0
\end{equation}

\begin{equation}
\resizebox{0.8\linewidth}{!}{
    $
    \begin{aligned}
    2\sum_{j=1}^{b_s-1}\cos\left(\dfrac{mj\pi}{b_s}\right)=\sum_{j=1}^{b_s-1}\left(\cos\left(\dfrac{mj\pi}{b_s}\right)+\cos\left(\dfrac{m\left(b_s-j\right)\pi}{b_s}\right)\right)=0
    \end{aligned}
    $}
\end{equation}

If $m$ is even, then

\begin{equation}
\resizebox{0.7\linewidth}{!}{
    $
    \begin{aligned}
        \sum_{j=1}^{b_s-1}\cos\left(\dfrac{mj\pi}{b_s}\right) &=\dfrac{1}{2}\left(\frac{\sin\left(\left(b_s-\frac{1}{2}\right)\dfrac{m\pi}{b_s}\right)}{\sin\left(\dfrac{m\pi}{2b_s}\right)}-1\right) \\ 
        &=\dfrac{1}{2}\left(\frac{\sin\left( m\pi-\dfrac{m\pi}{2b_s}\right)}{\sin\left(\dfrac{m\pi}{2b_s}\right)}-1\right)=-1
    \end{aligned}
    $}
\end{equation}

Since the count of odds in $1$ to $l$ is $\lfloor\frac{l+1}{2}\rfloor$ and the count of evens in $1$ to $l$ is $\lfloor\frac{l}{2}\rfloor$, so

\begin{equation}
    \sum_{m=1}^{l}\sum_{j=1}^{b_s-1}\cos\left(\dfrac{mj\pi}{b_s}\right)=\left\lfloor\frac{l+1}{2}\right\rfloor\times 0+\left\lfloor\frac{l}{2}\right\rfloor\times\left(-1\right)=-\left\lfloor\dfrac{l}{2}\right\rfloor
\end{equation}

\begin{equation}
\resizebox{0.8\linewidth}{!}{
    $
    \begin{aligned}
        \sum_{l=0}^{2k-1}\sum_{m=1}^{l}\sum_{j=1}^{b_s-1}\cos\left(\dfrac{mj\pi}{b_s}\right) &=-\sum_{l=0}^{2k-1}\left\lfloor\dfrac{l}{2}\right\rfloor = k-k^2
    \end{aligned}
    $}
\end{equation}

Finally,

\begin{equation}
\resizebox{0.8\linewidth}{!}{
    $
    \begin{aligned}
        \sum_{j=1}^{b_s-1}\dfrac{\sin^2\left(\frac{kj\pi}{b_s}\right)}{\sin^2\left(\frac{j\pi}{2b_s}\right)} &=2\sum_{j=1}^{b_s-1}\sum_{l=0}^{2k-1}\sum_{m=1}^{l}\cos\left(\dfrac{mj\pi}{b_s}\right)+2kb_s-2k \\
        &=2kb_s-2k^2
    \end{aligned}$
    }
\end{equation}

\noindent
\textbf{Analysis of $\Delta p$.}
According to the Central Limit Theorem, the sum of independent uniform variables approximates a normal distribution. Therefore, we can further analyze the error of each trajectory point in Equation~\eqref{eq:delta_p}, where $|p_k-p'_k|$ denotes the spatial distance between the original point $p_k\in \mathcal{T}$ and its reconstructed counterpart $p'_k\in \mathcal{T}'$.
\begin{equation}
    \Delta p_k = |p_k-p'_k| =  \sqrt{\sum\nolimits_{i=0}^{dim-1} \Delta S_{i,k}^2}
    \label{eq:delta_p}
\end{equation}

We can then calculate this distribution's Cumulative Distribution Function (CDF) to provide an appropriate method for selecting a proper $\epsilon_f$.
Each $S_{i,k}$ is independently, normally distributed with mean of $0$ and variance of $\text{Var}(\Delta S_{k})$, so $\Delta p_k$ follows a scaled chi-square distribution with $dim$ degrees of freedom.
Let $\sigma_k = \sqrt{\text{Var}(\Delta S_{k})} = \sqrt{\frac{(kb_s-k^2)\epsilon_f^2}{24b_s^2}}$. Then, the expected value and CDF of $\Delta p_k$ are:
\begin{equation}
    E\left(\Delta p_k\right) = \sigma_k \sqrt{2} \frac{\Gamma\left(\frac{dim+1}{2}\right)}{\Gamma\left(\frac{dim}{2}\right)}
\end{equation}
\begin{equation}
\resizebox{0.85\linewidth}{!}{
    $
    F_{\Delta p_k}\left(r\right) = \left\{
    \begin{aligned}
    2\Phi\left(\frac{r}{\sigma_k}\right) - 1 - \sqrt{\frac{2}{\pi}}\exp\left(-\frac{r^2}{\sigma_k^2}\right)\sum_{k=0}^{m-1}{\frac{1}{\left(2k+1\right)!!}\left(\frac{r^2}{\sigma_k^2}\right)^k} & \\
    , dim=2m+1 & \\
    1 - \exp\left(-\frac{r^2}{2\sigma_k^2}\right)\sum_{k=0}^{m-1}{\frac{1}{k!}\left(\frac{r^2}{2\sigma_k^2}\right)^k}, \quad dim=2m & \\
    \end{aligned}
    \right.
    $}
\end{equation}

Consider the point with the maximum expected error, at $p_{\frac{b_s}{2}}$. For $dim=2$, its CDF becomes
\begin{equation}
\resizebox{0.85\linewidth}{!}{
    $
    \begin{aligned}
        F_{\Delta p_{\frac{b_s}{2}}}\left(r\right) &= 1 - \exp\left(-\frac{r^2}{2\text{Var}\left(\Delta S_{\frac{b_s}{2}}\right)}\right) 
        = 1 - \exp\left(-\frac{12r^2}{\epsilon_f^2}\right)
    \end{aligned}
    $}
\end{equation}

The probability of exceeding a given maximum error limit $\epsilon$ is 
\begin{equation}
    P\left(\Delta p_{\frac{b_s}{2}} > \epsilon\right) = 1 -  F_{\Delta p_{\frac{b_s}{2}}}\left(\epsilon\right) = \exp\left(-\frac{12\epsilon^2}{\epsilon_f^2}\right)
\end{equation}

If we choose $\epsilon_f=\frac{\epsilon}{0.6}$, then this probability is 1.3\%, indicating that very few points on the trajectory will exceed the maximum error. If $\epsilon_f=\frac{\epsilon}{0.8}$, the percentage could be further reduced to 0.05\%.

We can also estimate the average trajectory error as:
\begin{equation}
\resizebox{0.85\linewidth}{!}{
    $
    \begin{aligned}
        E\left(\overline{\Delta T}\right) &= \sum_{k=0}^{b_s} E\left(\Delta p_k\right) = \sqrt{2} \frac{\Gamma\left(\frac{3}{2}\right)}{\Gamma\left(1\right)} \sum_{k=0}^{b_s}{\sigma_k} \\
        &= \sqrt{2} \frac{\frac{\sqrt{\pi}}{2}}{1} \sum_{k=0}^{b_s}{\sqrt{\frac{\left(kb_s-k^2\right)\epsilon_f^2}{6b_s^2}}} 
        = \frac{\sqrt{\pi}\epsilon_f}{\sqrt{12}} \sum_{k=0}^{b_s}{\sqrt{\frac{kb_s-k^2}{b_s^2}}}
    \end{aligned}
    $}
\end{equation}

Since $b_s$ is large, 
\begin{equation}
    \sum_{k=0}^{b_s}{\sqrt{\frac{kb_s-k^2}{b_s^2}}} \simeq \int_0^1{\sqrt{x-x^2}} = \frac{\pi}{8}
\end{equation}

Substituting it into the above equation, we can obtain: 
\begin{equation}
\resizebox{0.9\linewidth}{!}{
    $
    \begin{aligned}
        E\left(\overline{\Delta T}\right) &= \frac{\sqrt{\pi}\epsilon_f}{\sqrt{48}} \sum_{k=0}^{b_s}{\sqrt{\frac{kb_s-k^2}{b_s^2}}} 
        \simeq\frac{\sqrt{\pi}\epsilon_f}{\sqrt{12}} \frac{\pi}{8} 
        =\frac{\epsilon}{0.6} \frac{\pi\sqrt{\pi}}{8\sqrt{12}} \simeq 0.335\epsilon
    \end{aligned}\label{eq:avg_error}
    $}
\end{equation}

When $dim=3$, we can obtain similarly that $E(\overline{\Delta T}) \simeq 0.426\epsilon$.

%

\vspace{0.02in}
\noindent
\textbf{Complexity analysis.}
For uniform sampling data, the complexity of our model is $O(n\log(b_s))$, as the trajectory is partitioned into blocks of length $b_s$, with DCT and IDCT applied to each block. All other operations are linear in $n$. In practice, since $b_s$ is a constant for one trajectory, the overall complexity of our model is $O(n)$ where $n=|\mathcal{T}|$ is the total number of trajectory points. 
 
\subsection{Truncation and post-compression validation}
\label{subsec:validation}

Through our empirical study, we observe that the frequency arrays $\mathcal{F}_{i,j}$ produced by DCT often contain a large number of zeros. This behavior is attributed to DCT's energy compaction property, which concentrates most of the input sequence's energy (e.g., essential information) in the first few low-frequency coefficients. This effect is particularly evident in smooth or slowly varying sequences, such as the velocity sequences generated by our model. 
To further reduce storage requirements, we retain only a subset of the frequency components, specifically, the first $k$ low-frequency values, since higher-frequency components often have negligible impact on trajectory shape. 

Formally, given a frequency array $\mathcal{C}$ output by DCT, we retain only the first $k=b_s\times r_{ret}$ elements out of the total $b_s$ coefficients, where $r_{ret}$ is the retention rate (as defined in Equation~\eqref{equation:parameters}). The resulting array $\mathcal{F}$ is then given by: $\mathcal{F}=(F_1, F_2, \cdots, F_{k-1})$, where $F_i=\text{encode}(\text{round}(C_i, 2\epsilon_f))$ for $i\in [1, k-1]$. Recall $F_0=C_0=0$ is not stored in $\mathcal{F}$.
%
While this truncation strategy significantly reduces storage requirements, it introduces approximation errors and does not inherently guarantee that the reconstructed trajectory adheres to the specified error bound $\epsilon$. To address this limitation, we introduce a post-compression validation step. This step effectively ensures compliance with the error constraint and corrects any instances where the difference between the reconstructed and original positions at certain trajectory points exceeds $\epsilon$. 

In this step, the compressed representation $\cup\mathcal{F}_{i,j}$ is decompressed to reconstruct the trajectory $\mathcal{T}'$. We then compare $\mathcal{T}'$ with the original trajectory $\mathcal{T}$ to identify any points $p_i\in \mathcal{T}$ for which the reconstruction error exceeds the specified threshold $\epsilon$. The decompression procedure is detailed in Section~\ref{subsec:decompression}. 
The set of such error-prone points is defined as: $\mathcal{P}_e=\{p_i\in \mathcal{T}||p_i-p'_i|> \epsilon \}$. 

For each point $p\in \mathcal{P}_e$, we compute the dimension-wise differences between the original and reconstructed signals: $\cup_{i=0}^{dim-1} \Delta{S_{i,j}}=S_{i,j}-S'_{i,j}$. These differences are then scaled and quantized as: $\text{round}({\Delta S_{i,j}},{2\epsilon_{d}})$, where $\epsilon_{d}=\frac{\epsilon_p}{\sqrt{dim}}$ and $\epsilon_p$ denotes the point-wise error threshold. The quantized correction values are encoded using the enhanced Zigzag encoding, and the resulting correction data are aggregated into a correction array $\mathcal{A}_c$, defined as $\mathcal{A}_c=\{\cup_{i=0}^{dim-1}\Delta S_{i,j} \land t_j|p_j\in \mathcal{P}_e\}$. 
%
This correction array is stored as part of the additional metadata alongside the compressed trajectories. 
Although this validation step increases execution time due to the additional decompression and validation step, it effectively ensures that the final output meets the desired error tolerance.

\vspace{-0.08in}
\subsection{Storage structure}
\label{subset:storage}

The final step in the compression process involves storing all previously processed data. A visualization of the data storage structure is shown in Figure~\ref{fig:adjusted_storage_structure}. Note that the structure will be adjusted to support trajectory data with non-uniform sampling intervals (see Section~\ref{subsec:extension} for details). Consequently, we present a generalized version capable of handling arbitrary input trajectories. 
For each input trajectory $\mathcal{T}$, our compression model outputs the compressed frequency arrays $\cup_{i<dim, j<\lceil\frac{|\mathcal{T}|-1}{b_s}\rceil}\mathcal{F}_{i,j}$, along with following metadata. 

\vspace{0.02in}
\noindent
\textbf{Trajectory-level metadata.} The following metadata are associated with each trajectory $\mathcal{T}$: (1) $\Delta t$ - the sampling interval of the trajectory, used to reconstruct timestamps; 
(2) $\epsilon$ - the error bound used to compress the trajectory. This is needed both to recover values from rounded integers and to compute parameters such as $\epsilon_f$, $b_s$, and $r_{ret}$;
(3) $\epsilon_t$ - the time precision to quantize timestamps. 
(4) $|\mathcal{T}|$ - the total number of samples (i.e., the length) of the trajectory; 
(5) $t_0$ and $p_0$ - the initial timestamp and starting position of the trajectory;
and (6) Correction array $\mathcal{A}_c$, which stores information about error points: the total number of error points $|\mathcal{P}_e|$, and the set of timestamp-difference pairs for each dimension at every error point $\mathcal{A}_c=\{\cup_{i=0}^{dim-1}\Delta S_{i,j} \land t_j|p_j\in \mathcal{P}_e\}$.

\vspace{0.02in}
\noindent
\textbf{Block-level metadata.} The following metadata are associated with each $\mathcal{F}_{i,j}$: (1) Ending value of each block. For each segmented signal sequence $\mathcal{S}_{i,j}$, we store its ending value, which is also the starting value of the successor sequence $\mathcal{S}_{i,j+1}$. For example, in the $x$-dimension,  the first block's ending value is $S_{b_s}$, the second is $S_{2b_s}$, and so on. Using these, the average velocity $v_{avg}$ of each segmented signal sequence $\mathcal{S}_{i,j}$ can be derived as $\frac{S_{(j+1)b_s}-S_{jb_s}}{b_s}$. This reflects the overall trend within the block. To save space, we do not store the absolute values of $S_{i\cdot b_s}$ directly. Instead, we store the differences between consecutive block ending values (e.g., $S_{b_s}-S_{0}$, $S_{2\cdot b_s}-S_{b_s}$, etc). Any ending value $S_{i\cdot b_s}$ can be reconstructed using $p_0$ and the cumulative sum of these differences: $\sum_{j=0}^{i-1}(S_{(j+1)b_s}-S_{jb_s})$. 
(2) Number of frequency components $c_F$. Each truncated frequency array $\mathcal{F}_{i,j}$ may still include some trailing zeros. We further reduce storage by omitting those zeros. The counter $c_F$ associated with $\mathcal{F}_{i,j}$ records the number of meaningful components retained in $\mathcal{F}_{i,j}$. 

\vspace{0.02in}
\noindent
\textbf{Quantization and encoding.} To further reduce the storage cost, we apply quantization and encoding techniques, as explained in Section~\ref{subsec:point-encoding}. For floating numbers, we convert them to signed integer via quantization. Examples include: $\text{round}(\Delta S_{i,j}, 2\epsilon_d)$ where $\epsilon_d=\frac{\epsilon_p}{\sqrt{dim}}$, $\text{round}(p_0, 2\epsilon_p)$, $\text{round}(S_{(j+1)\cdot b_s}-S_{j\cdot b_s}, 2\epsilon_d)$, and $\text{round}(t_0, \epsilon_t)$.
Those signed integers are encoded via Zigzag encoding.
Unsigned integers (e.g., $|\mathcal{T}|$, $\cup c_F$, and $|\mathcal{P}_e|$) are encoded using Varint.

\subsection{Decompression process}
\label{subsec:decompression}

Next, we describe \model's decompression process. After loading all the data according to the previously defined storage structure, we  reconstruct the frequency array $\mathcal{C}'_{i,j}$. This is done by prepending $C'_0=0$ and padding ($b_s-c_F-1$) zeros to each output frequency array $\mathcal{F}_{i,j}$ to match the original block size $b_s$.
We then reconstruct the velocity array $\mathcal{V}'_{i,j}$ by applying ICDT to $\mathcal{C}'_{i,j}$. Using $\mathcal{V}'_{i,j}$ and the corresponding average velocity $v_{avg}$, which can be derived from the corresponding ending positions $S_{(j+1)b_s}$ and $S_{jb_s}$ along the $i$-th dimension and $p_0$, we can restore the signal array $\mathcal{S}'_{i,j}$. Finally, we recover the error points using the correction array $\mathcal{A}_c$.


\subsection{Extension to non-uniform sampling data}
\label{subsec:extension}

As previously mentioned, trajectory data collected from real-world applications often exhibit non-uniform sampling frequencies, requiring specific adaptations to our model for effective processing. One specific challenge is the presence of discontinuous sub-trajectories, when the temporal or spatial gap between two consecutive trajectory points is unusually large. 


\begin{figure}[h]
    \centering
    \includegraphics[width=\linewidth]{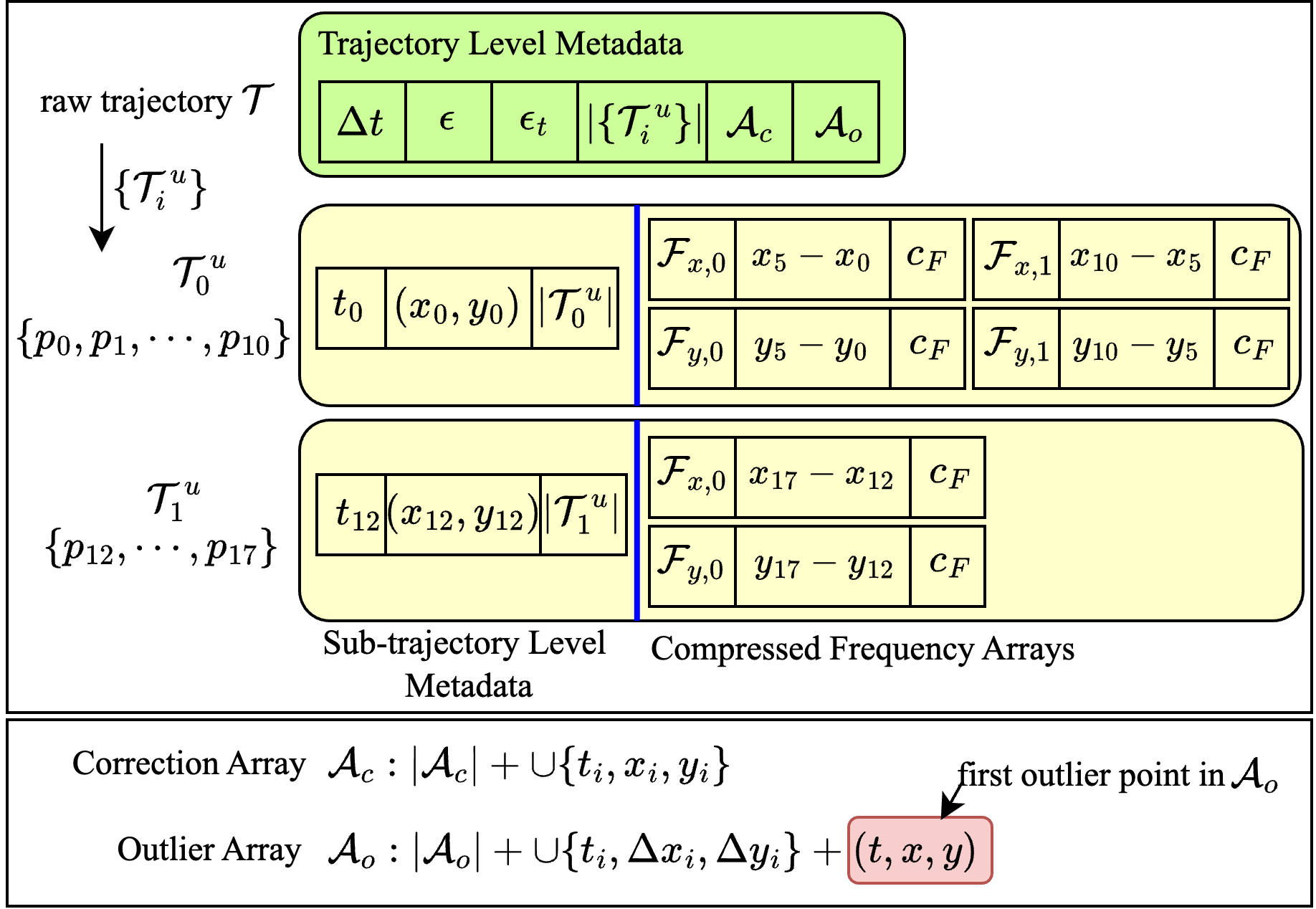}
    \vspace{-0.3in}
    \caption{The adjusted storage structure of \model\ ($b_s=5$, $dim=2$). 
    }
\label{fig:adjusted_storage_structure}
\vspace{-0.2in}
\end{figure}

To address this, we segment a raw input trajectory $\mathcal{T}$ into a set of temporally
continuous sub-trajectories, denoted as $\{ \mathcal{T}_i\}$. For each sub-trajectory $\mathcal{T}_i$, we then construct a uniformly sampled trajectory $\mathcal{T}^u_i$ and compress $\mathcal{T}^u_i$ via our \model. 

Since our model relies on the temporal continuity of trajectories and may experience degraded  performance when encountering noisy start or ending points, we begin by segmenting each input trajectory $\mathcal{T}$ into trajectory fragments \{$\mathcal{T}_i$\}, ensuring that all fragments are temporally continuous. This segmentation is based on the pointwise spatial and temporal distance between  consecutive trajectory points. Specifically, we use a predefined velocity threshold $v_{max}$ (e.g., $v_{max} = 200$ m/s for road vehicles),
and a time-based condition that reflects the expected sampling interval. A new segment $\mathcal{T}_{i+1}$ is initiated at point $p_j$ if either of the following conditions holds: $|p_j-p_{j-1}|>(t_j-t_{j-1}) \cdot v_{max}$ or $t_j-t_{j-1}> b_s \cdot \frac{t_j-\mathcal{T}_{i}.t_o}{|\mathcal{T}_{i}|}$, 
where $\mathcal{T}_{i}.t_o$ refers to the starting timestamp of the current trajectory fragment $\mathcal{T}_i$, and $b_s$ is the block size that allows for tolerance to moderate irregularities. In the edge case where $|\mathcal{T}_{i}| = 1$, we replace $\frac{t_j-\mathcal{T}_{i}.t_o}{|\mathcal{T}_{i}|}$ with a default value $\Delta t$ to handle the minimal-length fragment scenario. 
After segmentation, we assume, without loss of generality, that all trajectories $\{\mathcal{T}_i\}$ are temporally continuous.

We then determine the desired sampling interval $\Delta t$, which is set to the rounded average time gap between consecutive points across all temporally continuous sub-trajectories $\{\mathcal{T}_i\}$, i.e., $\Delta t= \left\lfloor\frac{\sum_{i}\mathcal{T}_i.t_{|\mathcal{T}_i|-1}-\mathcal{T}_i.t_0}{\sum_i|\mathcal{T}_i|}\right\rceil$. 
The choice of $\Delta t$ introduces a trade-off in   compression performance. A larger $\Delta t$ results in fewer samples in the uniformly sampled trajectory $\mathcal{T}^u_i$, improving the compression ratio. However, it also increases misalignment between $\mathcal{T}^u_i$ and original $\mathcal{T}_i$, which may require storing more correction information to keep the reconstruction error within the specified error bound $\epsilon$. In constrast, a smaller $\Delta t$ better preserves fine-grained details from the original trajectory but increases the length of $\mathcal{T}^u_i$, and thus the storage cost. Although one could search for an optimal $\Delta t$ by evaluating multiple values, we find that our \model\ delivers strong performance even with this simple average-based setting of $\Delta t$.

Next, we construct a uniformly sampled version $\mathcal{T}_i^u$ for each temporally continuous sub-trajectory $\mathcal{T}_i$, denoted as $\mathcal{T}_i^u=\langle(p_0,t_0)$, $(p^u_1, t_0+\Delta t), (p^u_2, t_0+2\Delta t), \cdots \rangle$,
where $(p_0,t_0)$ is the first point in the original trajectory $\mathcal{T}_i$.  
We then generate points $p^u_j\in \mathcal{T}^u_i$ at timestamp ($t_0+j\Delta t$) using linear interpolation. More specifically, let $(p_a, t_a)$ and $(p_{a+1}, t_{a+1})$ be two consecutive points in the original trajectory $\mathcal{T}_i$ such that $(t_0+j\Delta t) \in (t_a, t_{a+1}]$. The $x$-coordinate of $p^u_j\in \mathcal{T}^u_i$ is then approximated as $p^u_j.x = \frac{p_{a+1}.x-p_a.x}{t_{a+1}-t_a}\cdot (t_0+j\Delta t - t_a)+p_a.x$. Coordinate(s) in other spatial dimension(s) are interpolated in the same manner.

Finally, we make a few key adjustments to support this process: (1) Outlier Handling: Extremely short sub-trajectories (e.g., of length 1 or 2) are treated as outliers. The corresponding points (w.r.t $\mathcal{T}$) are stored in an outlier array $\mathcal{A}_o$, sorted by their timestamps. Each point $p\in \mathcal{A}_o$ is stored as a set of signed integers representing the difference from its predecessor outlier point along each spatial dimension:  $\cup_{i<dim}\text{round}(\Delta S, \frac{2\epsilon}{\sqrt{dim}})$, where $\Delta S$ indicates the difference between $p$ and the predecessor outlier along one spatial dimension. Those signed integers are encoded via Zigzag encoding to reduce storage overhead. Similarly, timestamps are converted to unsigned integer via $\text{round}(t, \epsilon_t)$ and then encoded via Varint encoding. 
(2) Modified storage structure: The storage structure is adapted to support multiple sub-trajectories. Since each sub-trajectory $\mathcal{T}^u_i$ is independent, each $\mathcal{T}^u_i$ stores its own trajectory-level metadata, including start time $t_0$, start point $p_0$, and its length $|\mathcal{T}^u_i|$. Other trajectory-level metadata, including $\Delta t$, $\epsilon$, $\epsilon_t$, the number of sub-trajectories $|\{\mathcal{T}_i^u\}|$, outlier array $\mathcal{A}_o$, and correction array $\mathcal{A}_c$ are associated with original trajectory $\mathcal{T}$.
A schematic of the updated storage structure is shown in Figure~\ref{fig:adjusted_storage_structure}. 
(3) Timestamp encoding: Though all individual timestamps $\{t_j\}$ are treated as ground truth 
when decompressing the trajectory, we need to store the exact timestamps of $\mathcal{A}_o$ and $\mathcal{A}_c$. Also, for other line simplification algorithms, we also store the exact timestamps for each reserved trajectory points.
Given such a sequence of timestamps, we first quantize each $t_j$ via $\text{round}(t_j, \epsilon_t)$. We then compute the differences between successive quantized timestamps and encode these differences - positive integers - using Varint encoding.
(4) Decompression adjustments: The decompression pipeline is modified accordingly. For each $\cup \mathcal{F}_{a,b}$ corresponding to a sub-trajectory $\mathcal{T}_i$, we first reconstruct a uniformly sampled sub-trajectory $(\mathcal{T}^u_i)'$ using the decompression process presented in Section~\ref{subsec:decompression}. 
Next, we reconstruct $\mathcal{T}'_i$ by aligning $(\mathcal{T}^u_i)'$ with the original timestamps $\{t_j\}$. To be more specific, for each timestamp $t_j$, we locate two consecutive points $(p^u_b)'$ and $(p^u_{b+1})'$ in $(\mathcal{T}^u_i)'$ such that $t_j\in (\mathcal{T}^u_i.t_0+b\Delta t, \mathcal{T}^u_i.t_0+(b+1)\Delta t]$. We then compute the interpolated position $p'_j$ using linear interpolation between these two points. We correct error points using the correction array $\mathcal{A}_c$. Finally, we concatenate the reconstructed sub-trajectories $\mathcal{T}'_i$s and apply outlier array $\mathcal{A}_o$ to represent the complete decompressed trajectory $\mathcal{T}'$.


\subsection{Strengths and limitations. }
To the best of our knowledge, \model\ is the first trajectory compression method to exploit the physical properties of trajectories by modeling them as continuous curves. This geometric perspective enables more accurate curve fitting and effectively reduces the average error. The error introduced by \model\ is mathematically bounded, making it well-suited for applications that require strict accuracy guarantees. \model\ operates in $O(n)$ time, independent of dimensionality, ensuring strong scalability and low computational overhead even in high-dimensional data spaces. This makes \model\ applicable not only to conventional 2D or 3D spatial data, but also to complex, multi-dimensional trajectory datasets found in domains like biomechanics, climate modeling, and robotics. However, like other SED-based methods, \model\ has limitations. It relies on timestamp information, and thus cannot handle trajectories lacking temporal data. It is also not directly compatible with other distance metrics such as PED. Furthermore, \model\ performs best when trajectories have relatively stable sampling intervals. Its accuracy and compression effectiveness degrade when facing highly irregular sampling patterns (e.g., intervals alternating between 1 and 10 seconds), as this distorts the curve-fitting process and reduces fidelity.


\begin{table*}[]
\vspace{-0.1in}
    \centering
    \caption{Real-life trajectory datasets}
    \vspace{-0.15in}
    \resizebox{1\linewidth}{!}{
    \begin{tabular}{|c|c|c|c|c|c|c|c|}
    
    \hline
    Datasets & Tra. Count & Sampling Intervals (s) & Time precision $\epsilon_t$ (s)  & $\Delta t$ (s) & Var($\Delta t$) &Avg tra. length & $a, b, c, d$ \\
    \hline
    nuPlan & 1,349 & 0.1  & 0.01 & 0.1 & $1.25\times {10^{-8}}$ & ~2,660 & 0.6, 20, 100, 0.04\\
    \hline
    GeoLife/GeoLife-3D & 182 & 1-5 & 1 & 3 & 18.95 & ~132,970 &0.6/0.7, 0.5, 25, 1.1/0.8 \\
    \hline
    Mopsi & 51 & 2  & 0.001  & 2 & 5.18 & ~149,020  & 0.6, 1, 25, 0.6\\
    \hline
    \end{tabular}}
    \label{tab:datasets}
    \vspace{-0.2in}
\end{table*}

\section{Experimental Study}
\label{sec:experimental_study}

In this section, we present a comprehensive experimental evaluation of our \model\ algorithm in comparison to state-of-the-art line simplification methods. Using four real-world trajectory datasets, we conducted a series of experiments to address the following aspects:
(1) The effectiveness of various strategies proposed in our paper, including point quantization and encoding technique introduced in Section~\ref{subsec:point-encoding}, blocking strategy in Section~\ref{subsec:framework}, 
and truncation strategy presented in Section~\ref{subsec:validation}. 
(2) The compression ratios achieved by \model\ compared to baselines. (3) The average error associated with each method. (4) Their performance on 3D trajectory datasets.

\vspace{-0.05in}
\subsection{Experimental settings}

\subsubsection{Real-life trajectory datasets} To align with our assumption of uniform sampling, we first use nuPlan~\cite{caesar2021nuplan}, a widely used benchmark in the autonomous driving domain. 
Specifically, we use one subset (i.e., \textbf{ego car trajectories}) from the test dataset, as only the ego car has a complete trajectory throughout the entire scene. This subset contains over 1,000 trajectories, each with approximately 2,660 trajectory points. It is important to note that for line simplification tasks, the number of points per trajectory plays a more critical role than the total number of trajectories in the dataset. This is because simplification quality and compression gains are primarily influenced by the density and continuity of points within individual trajectories. 

To demonstrate the adaptability of our model to non-uniformly sampled datasets, we further evaluate it on GeoLife~\cite{zheng2010geolife} and Mopsi~\cite{mariescu2017grid}. GeoLife comprises GPS trajectory collected from 182 users with varying sampling rates, while Mopsi includes GPS trajectories from 51 users sampled at 2-second intervals. Both datasets are mainly formed by long trajectories, with each on average containing more than 130,000 trajectory points. 

To evaluate the effectiveness of our model on 3D trajectory data, we enhance GeoLife by adding altitude information, resulting in a derived dataset referred to as GeoLife-3D. Although Mopsi also contains altitude information, it contains a high proportion of missing values, making it less suitable for 3D evaluation. A summary of the four datasets is provided in Table~\ref{tab:datasets}. Time precision is determined based on the original dataset (e.g., Mopsi stores timestamps as integer milliseconds, while GeoLife stores them as integer seconds).  



\subsubsection{Data preprocess} 

We begin by removing all the trajectory points with duplicate timestamps. Since GeoLife and Mopsi datasets provide GPS data in latitude and longitude, it is inconvenient to compute and constraint the maximum SED error directly in geographic coordinates. To address this, we select an appropriate UTM zone for each dataset and project all trajectories accordingly. The projected trajectories serve as the ground truth in our experiments. 

For 3D trajectory, we convert (longitude, latitude, altitude) coordinates into Cartesian ($X$, $Y$, $Z$) space using the WGS84 ellipsoid model. Although GeoLife dataset has some missing altitude data, these are mostly confined to a single trajectory. Thus, we retain default values -GeoLife dataset has a default value of -777 feet for invalid altitude, and we follow this for the few missing points and construct GeoLife-3D accordingly. 

\subsubsection{Algorithms and implementation} We evaluate four algorithms in our experiments: \model\ proposed in this paper, two state-of-the-art SED-based line simplification algorithms CISED-S and CISED-W, and SQUISH, selected for its adaptability to 3D trajectories. CISED-S and CISED-W are both one-pass, error-bounded line simplification algorithms. However, CISED-S is a strong simplification algorithm, meaning all points in the compressed trajectory are drawn from the original trajectory. In contrast, CISED-W is a weak simplification algorithm, which allows the inclusion of newly generated points not present in the original trajectory. 
All algorithms were implemented in Python, and experiments were conducted on a 16 Core AMD Ryzen 9 5950X @ 3.40GHz. 

\subsubsection{Evaluation metrics} We use the following metrics to evaluate algorithm performance.

\noindent
\textbf{Compression ratio.} Traditional trajectory simplification algorithms calculate compression rates as the ratio of retained points. Since our model uses transformed representations instead of points, we measure the compression rate based on the actual storage size. For a set of trajectories $\cup \mathcal{T}$ and their compressed version $\cup \mathcal{R}$, the compression ratio is defined as: $\sum\nolimits_{\mathcal{R}_i\in \cup \mathcal{R}}{\text{Size}\left(\mathcal{R}_i\right)} / \sum\nolimits_{\mathcal{T}_i\in \cup \mathcal{T}}{\text{Size}\left(\mathcal{T}_i\right)}$.
%
%
Here, $\text{Size}$ denotes the total byte size. For example, if each 2D point is stored as three double-type values ($x$, $y$, $t$), the storage for a trajectory $\mathcal{T}_i$ with $|\mathcal{T}_i|$ points is $192 \times |\mathcal{T}_i|$ bytes. Lower values indicate more effective compression. 

\noindent
\textbf{Max SED error.} The compressed trajectories of our data are not polygonal lines, so we cannot use the distance between line segments and points as the error metrics. Hence, error is defined as the maximum Euclidean distance between origin and reconstructed points: $\max_{\mathcal{T}_i\in \cup \mathcal{T}}\left({\max\nolimits_{p_j\in \mathcal{T}_i}{|p_{j},p'_{j}|}}\right)$.
%
%
Here, $p_{j}$ and $p'_{j}$ are corresponding to points in the original trajectory $\mathcal{T}_i$ and reconstructed trajectory $\mathcal{T}'_i$ respectively, and $|p,p'|$ is an operation to calculate the distance between two points. 

This max SED error serves as the overall error bound $\epsilon$. Due to varying sampling rates across datasets, we use different max error ranges. Specifically, we vary $\epsilon$ from $10$ to $100$ meters on datasets GeoLife, Mopsi, and GeoLife-3D, and vary $\epsilon$ from $1$ to $10$ meters on dataset nuPlan, since its sampling rate is only $0.1s$ and is about one-tenth of the other datasets. 

\noindent
\textbf{Mean SED error.} Defined as the average distance between corresponding original and reconstructed points: \\
$\Big(\sum\nolimits_{\mathcal{T}_i \in \cup \mathcal{T}}{\sum\nolimits_{p_j\in \mathcal{T}_i|}{|p_{j},p'_{j}|}}\Big)/ \sum\nolimits_{\mathcal{T}_i\in \cup \mathcal{T}}{|\mathcal{T}_i|}$.
%

\subsubsection{Parameters} Our \model\ model includes three parameters: the maximum allowable frequency error $\epsilon_f$, block size $b_s$, and retention rate $r_{ret}$.
These parameters are all functions of $\epsilon$, as defined in Equation~\eqref{equation:parameters}. The constraint values $a$, $b$, $c$, and $d$ are dataset-specific and are empirically determined through experimentation, as listed in Table~\ref{tab:datasets}.
We evaluate the sensitivity of these parameters and their impact on performance through a series of experiments. 
\begin{equation}
\epsilon_f = \frac{\epsilon}{a}, \quad b_s = b * \epsilon + c, \quad r_{ret} = \min\left(1, \frac{d}{\sqrt{\epsilon}}\right)
\label{equation:parameters}
\end{equation}
%
%
%
%
%
%

CISED requires a single parameter $m$, which refers to the number of sides of the regular polygon. Following the research findings reported in~\cite{lin2019one}, we use $m=16$, as increasing beyond this yields negligible gains in compression rate (<1\%).

In addition to the parameters required by compression algorithms, we also introduce two additional encoding parameters, $l$ and $\epsilon_p$ when using encoding to further reduce the storage overhead. Specifically, $l$ refers to the length of the chunks in encoding methods Varint and Zigzag, and $\epsilon_p$ records the maximum tolerable error when rounding points to integers. The default value of $l$ is $2$ for all algorithms, while the default value of $\epsilon_p$ is $0.5\epsilon$ for \model\ and $0.2\epsilon$ for other algorithms. Note that our model utilizes $\epsilon_p$ in a way different from other algorithms. Consequently, they have different default value for $\epsilon_p$. The impact of these parameters is also examined in our experimental analysis.

\subsection{Experimental results}

\begin{figure*}[h]
    \centering
    \begin{subfigure}{\linewidth}
        \includegraphics[width=\linewidth]{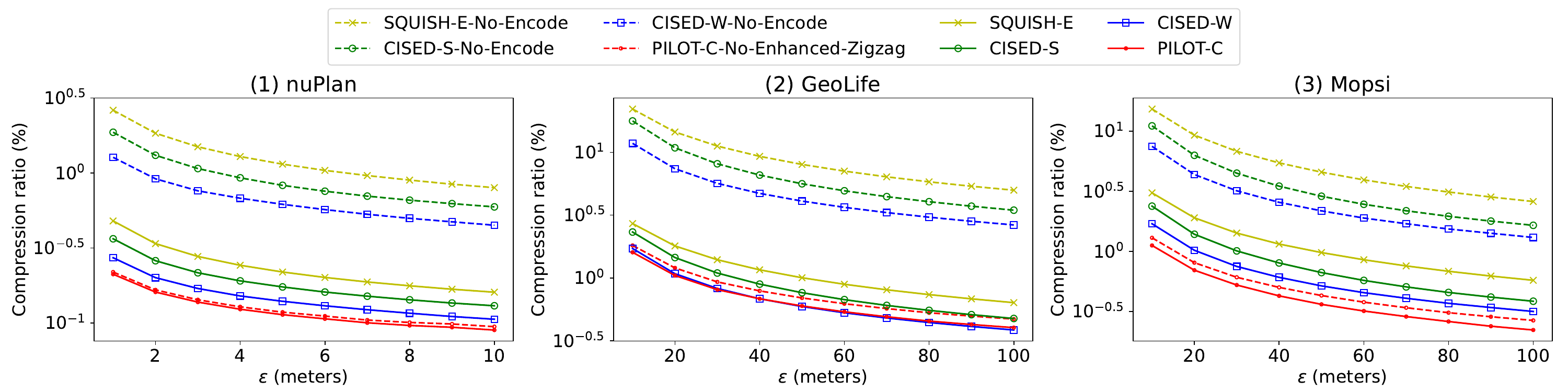}
    \end{subfigure}
    \vspace{-0.33in}
    \caption{Compression ratios vs. $\epsilon$.
    }
    \vspace{-0.1in}
    \label{fig:Exp-test_encode}
\end{figure*}

\begin{figure*}[h]
\vspace{-0.07in}
    \centering
    \begin{subfigure}{\linewidth}
        \includegraphics[width=\linewidth]{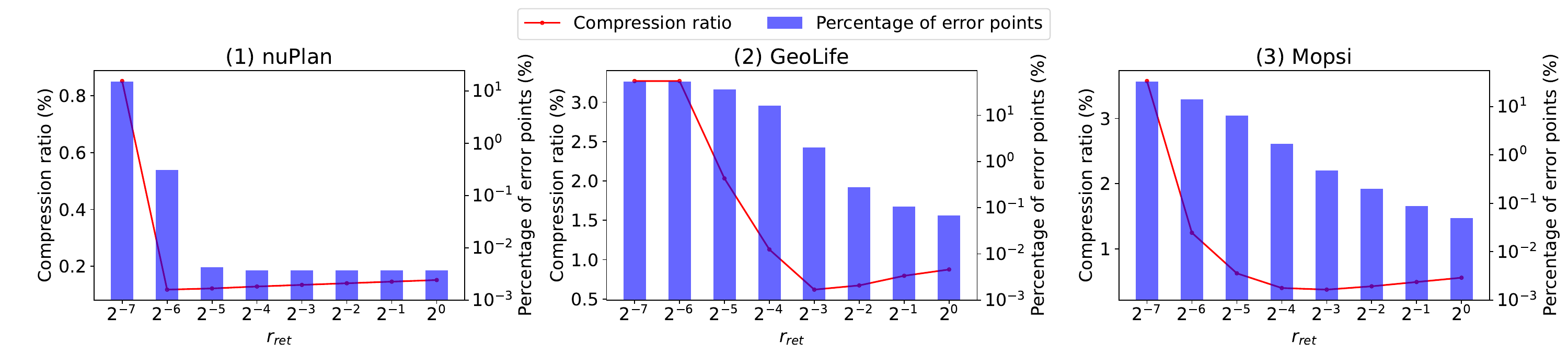}
    \end{subfigure}
    \vspace{-0.35in}
    \caption{Compression ratios and percentage of error points vs. $r_{ret}$.
    }
    \vspace{-0.1in}
    \label{fig:Exp-test_ret}
\end{figure*}
\begin{figure*}[h]
\vspace{-0.07in}
    \centering
    \begin{subfigure}{\linewidth}
        \includegraphics[width=\linewidth]{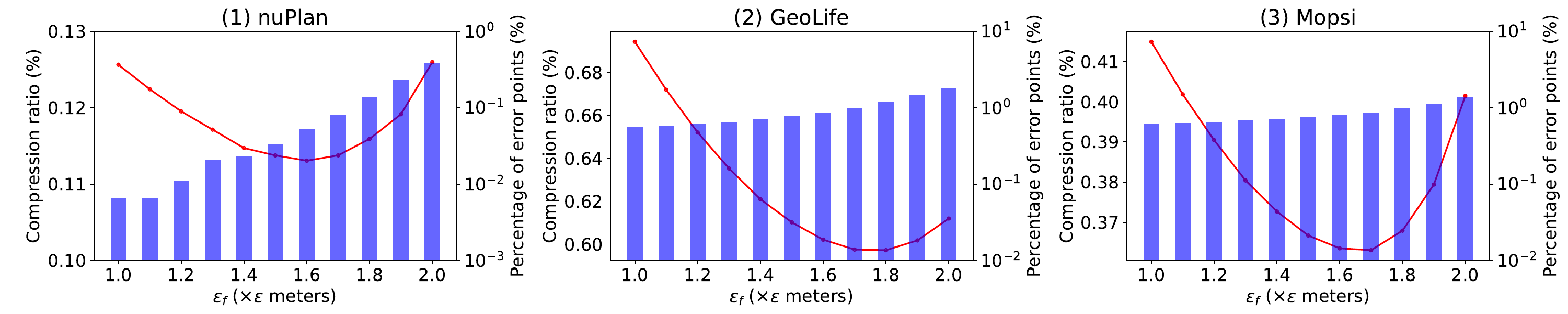}
    \end{subfigure}
    \vspace{-0.35in}
    \caption{Compression ratios and percentage of error points vs. $\epsilon_f$.
    }
    \vspace{-0.15in}
    \label{fig:Exp-test_epsilon-f}
\end{figure*}

\begin{figure}[h]
    \centering
    \begin{subfigure}{0.70 \linewidth}
        \includegraphics[width=\linewidth]{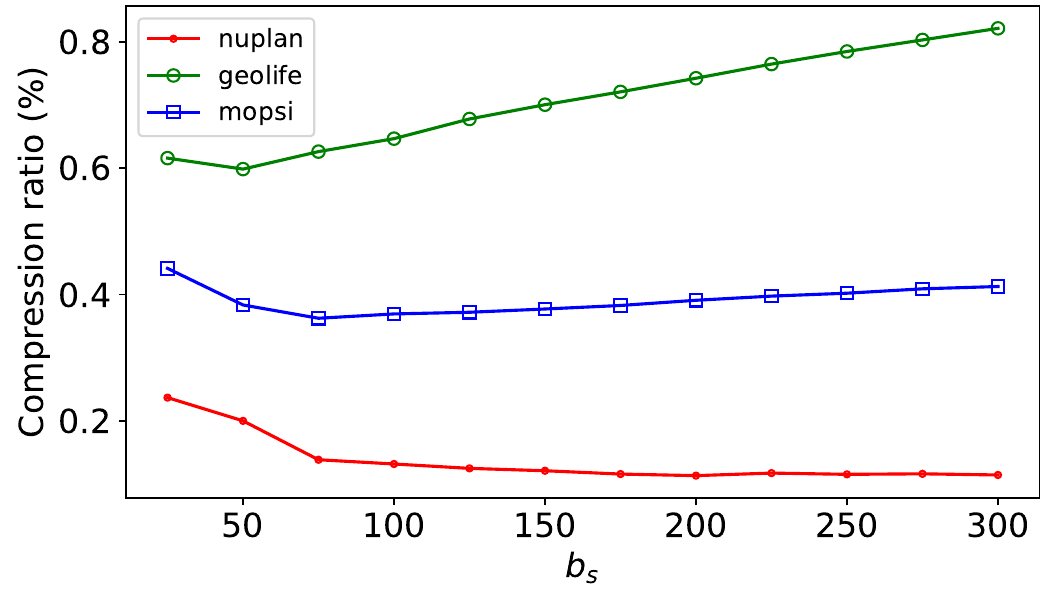}
    \end{subfigure}
    \vspace{-0.2in}
    \caption{Compression ratios of \model\ vs. $b_s$.
    }
    \vspace{-0.2in}
    \label{fig:Exp-test_b_s}
\end{figure}

\subsubsection{Effectiveness of point quantization/encoding, blocking, and truncation strategies}

We begin by evaluating the generality and effectiveness of the quantization and encoding techniques introduced in Section~\ref{subsec:point-encoding}, applying them to all three baseline algorithms. In Figure~\ref{fig:Exp-test_encode}, we compare the compression ratios of the original baseline methods (labeled ``No-Encode'') with their enhanced versions that incorporate these techniques, tested under different $\epsilon$ values while keeping other parameters to their default settings. The results clearly demonstrate that quantization and encoding consistently improve compression ratios across all datasets and $\epsilon$ values, with an average improvement of 455\%. This substantial improvement is primarily due to the high redundancy in storing compressed points using double-precision values. By reducing storage precision, the quantization and encoding methods significantly enhance compression efficiency. For the remainder of our experimental study, we only report the performance of the enhanced versions of the baseline methods, i.e., those augmented with quantization and encoding.

Next, we evaluate the impact of the enhanced Zigzag encoding technique, which is applied exclusively in  \model. As shown in Figure~\ref{fig:Exp-test_encode}, the enhanced Zigzag improves \model's compression performance by an average of 12.6\%. This modest improvement stems from the relatively minor advantage of the enhanced Zigzag encoding over the original Zigzag. 

\model\ partitions long sequences into blocks of fixed size. As discussed in Section~\ref{subsec:framework}, this segmentation  improves compression efficiency. To illustrate this effect, we report the compression ratio of \model\ under varying block sizes $b_s$ in Figure~\ref{fig:Exp-test_b_s}. As observed, increasing $b_s$ from small values initially improves compression ratio. However, beyond a certain point, further increases in $b_s$ lead to a decline in compression performance. The optimal value of $b_s$ - at which \model\ achieves the highest compression ratio - varies across datasets, as it is influenced by their respective sampling rates. While empirical searches can be used to find the optimal $b_s$, this process must be repeated whenever the error bound $\epsilon$ changes. In contrast, we set $b_s$ according to Equation~\eqref{equation:parameters}. Although the parameters $b$ and $c$ require empirical tuning, the equation provides a consistent and effective way to estimate good values for $b_s$ across different error bounds. As a result, \model\ maintains strong compression performance without re-tuning $b_s$ for every new setting. 

As discussed in Section~\ref{subsec:validation}, the frequency arrays $\mathcal{F}_{i,j}$ produced by DCT often contain many zeros, particularly in high-frequency coefficients. To improve compression, we apply truncation to retain only the first few low-frequency coefficients. To evaluate the impact of this strategy, we report the compression ratios and percentage of error points (exceed $\epsilon$) of \model\ under different retention rates in Figure~\ref{fig:Exp-test_ret}. Note that when $r_{ret}=2^0$, no elements are removed from the frequency arrays $\mathcal{F}_{i,j}$ produced by DCT, representing the upper extreme; when $r_{ret}=2^{-7}$, only the first 1 or 2 elements of each array are retained, resulting in the loss of important information representing the lower extreme. As expected, reducing the retention rate from $2^0$ initially improves the compression ratio, while the percentage of error points remains nearly unchanged. This indicates that removing high-frequency coefficients does not significantly impact the preservation of important information. However, when the retention rate becomes too low ($2^{-7}$ or $2^{-6}$), critical information is lost from the compressed frequency arrays. This leads to a noticeable increase in error points, which in turn increases the size of the correction array $\mathcal{A}_c$, ultimately reducing overall compression efficiency. Consequently, selecting an appropriate retention rate is crucial to maintaining both compression efficiency and data fidelity. While it is possible to empirically determine an optimal $r_{ret}$ through testing, our method, as presented in Equation\eqref{equation:parameters}, provides a simple and effective way to estimate a suitable $r_{ret}$ value.


\vspace{-0.05in}
\subsubsection{Compression performance}

In our second set of experiments, we compare the compression performance of \model\ against baseline algorithms. For each experiment, we vary a single parameter while keeping all other parameters fixed at their default values.

\noindent
\textbf{Compression ratios vs. error bound $\epsilon$.} We first evaluate how compression performance varies with the error bound. On the GeoLife and Mopsi datasets, we vary $\epsilon$ from $10$ to $100$ meters, and from $1$ to $10$ meters on nuPlan. The results are shown in Figure~\ref{fig:Exp-test_encode}; please focus on the solid lines. As $\epsilon$ increases, the compression ratios of all algorithms decrease across all datasets. Among the datasets, nuPlan has the best compression ratios, due to its high sampling rate. Across all tested $\epsilon$ values, \model\ consistently outperforms CISED-S and SQUISH-E on all datasets. It also outperforms CISED-W on nuPlan and Mopsi, and achieves comparable performance on GeoLife. Compared to nuPlan and Mopsi, GeoLife exhibits an irregular sampling pattern (i.e., $\Delta t=3$ seconds with Var($\Delta t$) close to 19), which affects the effectiveness of \model.  
On average, \model\ achieves compression ratios of (65.2\%, 78.1\%, 54.1\%) and (81.8\%, 100.3\%, 69.6\%) relative to CISED-S and CISED-W, respectively, across three datasets. 

\begin{figure*}[h]
\vspace{-0.1in}
    \centering
    \begin{subfigure}{\linewidth}
        \includegraphics[width=\linewidth]{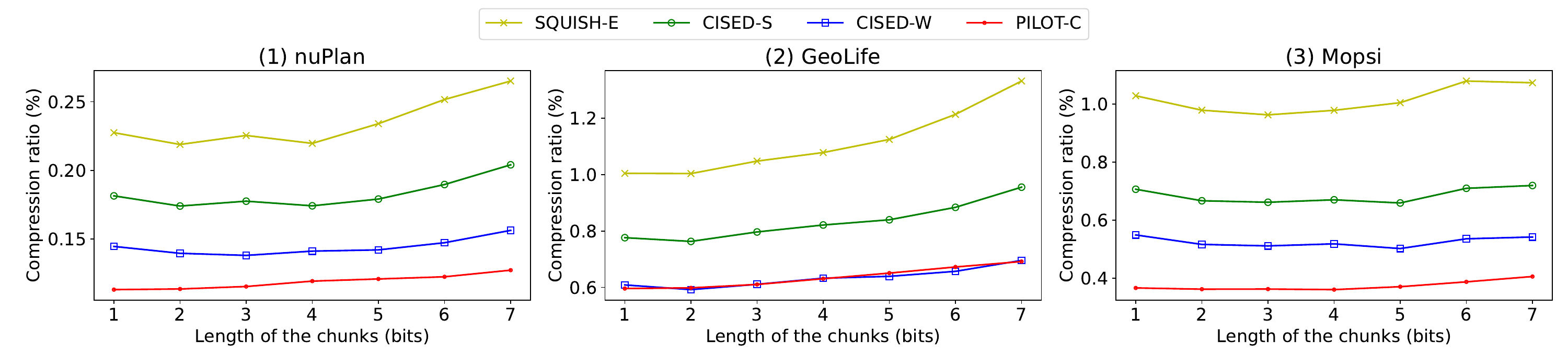}
    \end{subfigure}
    \vspace{-0.35in}
    \caption{Compression ratios vs. $l$. 
    }
    \vspace{-0.1in}
    \label{fig:Exp-test_utf}
\end{figure*}

\noindent
\textbf{Compression ratio vs. chunk length $l$.} To evaluate the impacts of the chunk length parameter $l$, which is used in the encoding phase of all algorithms, we fix the error bound $\epsilon=50$ m for the GeoLife and Mopsi, and $\epsilon=5$ m for nuPlan. We vary $l$ from $1$ to $7$ and the results are reported in Figure~\ref{fig:Exp-test_utf}. 
Across all algorithms, the compression ratio exhibits a non-monotonic trend: it initially decreases with increasing $l$, then increases beyond a certain point. This behavior stems from the variable effectiveness of Varint and Zigzag encoding, which are sensitive to chunk alignment and data distribution. 
%
The optimal $l$ value varies across datasets, particularly in Mopsi. This variation is partly due to differences in timestamp precision. Mopsi records timestamps with $0.001$-second resolution, but its average sampling interval is $2$ seconds, resulting in redundancy and inefficiencies in certain encoding configurations.
Given that (i) several $l$ values yield similar compression performance, and (ii) the optimal $l$ varies depending on $\epsilon$, we adopt a uniform default of $l=2$ for all algorithms in subsequent experiments to ensure consistency and simplicity.


\begin{figure*}[h]
\vspace{-.07in}
    \centering
    \begin{subfigure}{\linewidth}
    \includegraphics[width=\linewidth]{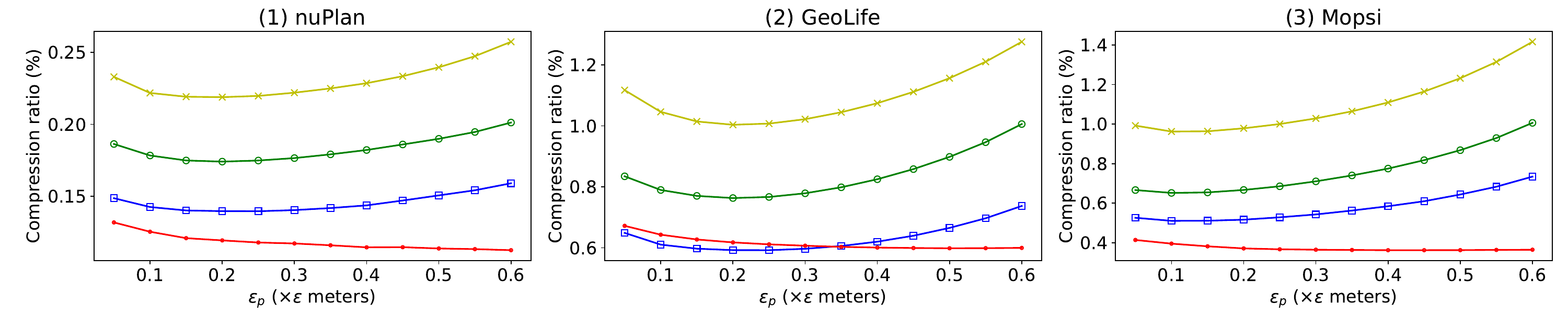}
    \end{subfigure}
    \vspace{-0.35in}
    \caption{Compression ratios vs. $\epsilon_p$.}
    \label{fig:Exp-test_accuracy}
    \vspace{-0.1in}
\end{figure*}

\begin{figure*}[h]
\vspace{-.07in}
    \centering
    \begin{subfigure}{\linewidth}
        \includegraphics[width=\linewidth]{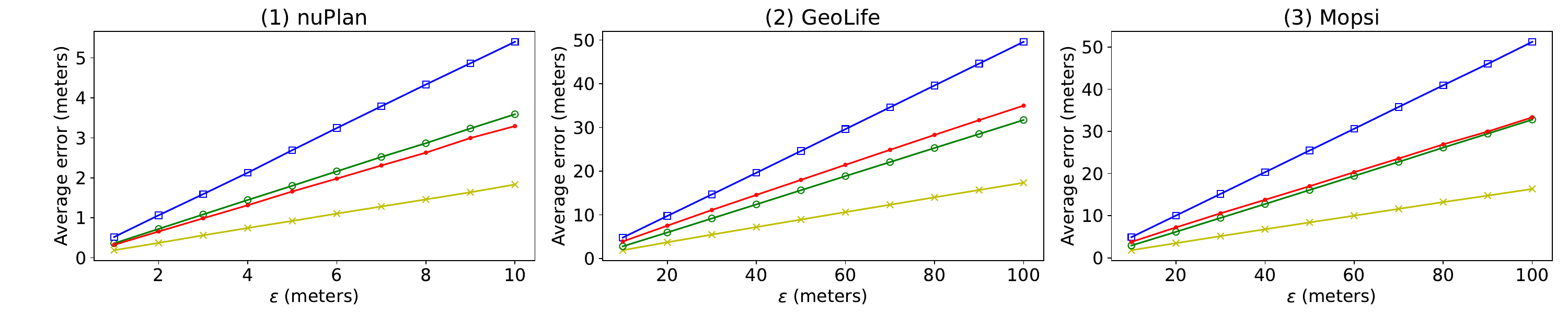}
    \end{subfigure}
    \vspace{-0.35in}
    \caption{Average errors vs. $\epsilon$. 
    }
    \vspace{-0.15in}
    \label{fig:Exp-test_average}
\end{figure*}

\noindent
\textbf{Compression ratio vs. frequency precision $\epsilon_f$.} As analyzed in Section~\ref{subsec:correctness}, the frequency precision $\epsilon_f$ directly affects the number of points on the trajectory that exceed the maximum allowable error $\epsilon$ and thus influence the size of the correction array. As $\epsilon_f$ increases, more error points emerge, as evidenced in Figure~\ref{fig:Exp-test_epsilon-f}. On the other hand, increasing $\epsilon_f$ also reduces the value of $\text{round}({C_i}, {2\epsilon_f})$, leading to shorter encoding lengths and lower storage requirements. Consequently,
selecting an appropriate $\epsilon_f$ is crucial. Since the optimal values are generally consistent across all datasets, we adopt the same values for all 2D datasets, as presented in Equation\eqref{equation:parameters} and Table~\ref{tab:datasets}, while 3D dataset has a different setting.

\noindent
\textbf{Compression ratio vs. point precision $\epsilon_p$.} Next, we examine the impacts of $\epsilon_p$, which controls the precision of stored points. We fix $\epsilon=50$ meters for GeoLife and Mopsi, and $\epsilon=5$ meters for nuPlan. The value of $\epsilon_p$ is varied from $ 0.05\times\epsilon$ to $0.6\times\epsilon$. Results are shown in Figure~\ref{fig:Exp-test_accuracy}.
As $\epsilon_p$ increases, the compression ratios of all algorithms first decrease and then increase. This reflects the trade-off between reduced storage precision and the ability of encoding methods to efficiently compress the data. The optimal value of $\epsilon_p$ that yields the best compression is relatively consistent across datasets but varies among algorithms. This is due to how each algorithm utilizes $\epsilon_p$. In \model, $\epsilon_p$ is only used during post-compression validation step to construct the correction array, leaving the core compression process unaffected. In contrast, CISED and SQUISH-E incorporate $\epsilon_p$ into their internal compression thresholds, directly influencing overall compression. 
%
%
Since multiple $\epsilon_p$ values often yield similar compression performance, we adopt fixed defaults for simplicity in subsequent experiments:  
$\epsilon_p = 0.5\epsilon$ for \model\ and $\epsilon_p = 0.2\epsilon$ for baseline algorithms.

\subsubsection{Average error comparison}

In the next set of experiments, we compare the average errors of different algorithms under varying $\epsilon$ and report the results in Figure~\ref{fig:Exp-test_average}.
%
%
As expected, the average errors increase with increasing $\epsilon$. More specifically, the average error of each algorithm shows an approximately linear relationship with $\epsilon$.
The average error of \model\ is close to that of CISED-S, smaller than that of CISED-W, and larger than that of SQUISH-E. \model\ outperforms CISED-W because it incorporates the physical characteristics of the trajectory and treats the trajectory as a continuous curve for fitting. In addition, the experimental results are consistent with the theoretical finding presented in Equation~\eqref{eq:avg_error}, i.e., the average error of \model\ is approximately $0.335\epsilon$ for 2D datasets. SQUISH-E produces the smallest errors, as it tends to retain more points, resulting in much weaker compression performance. 



\begin{figure}[h]
    \centering
    \begin{subfigure}{0.49 \linewidth}
        \includegraphics[width=\linewidth]{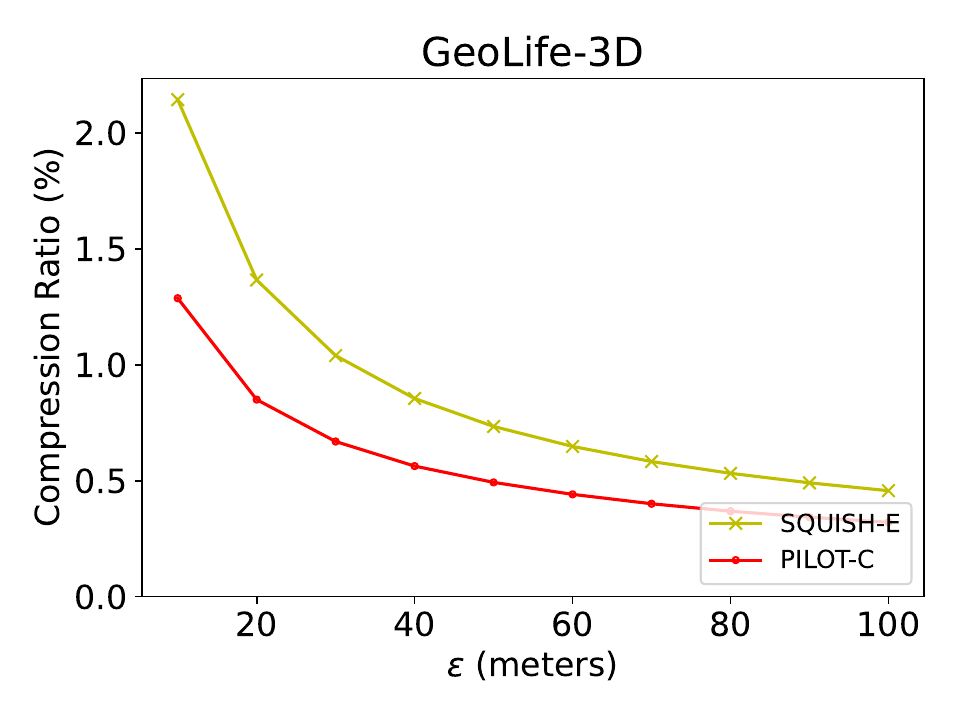}
        \vspace{-0.15in}
        \caption{Compression Ratio}
    \end{subfigure}
    \hfill
    \begin{subfigure}{0.49 \linewidth}
        \includegraphics[width=\linewidth]{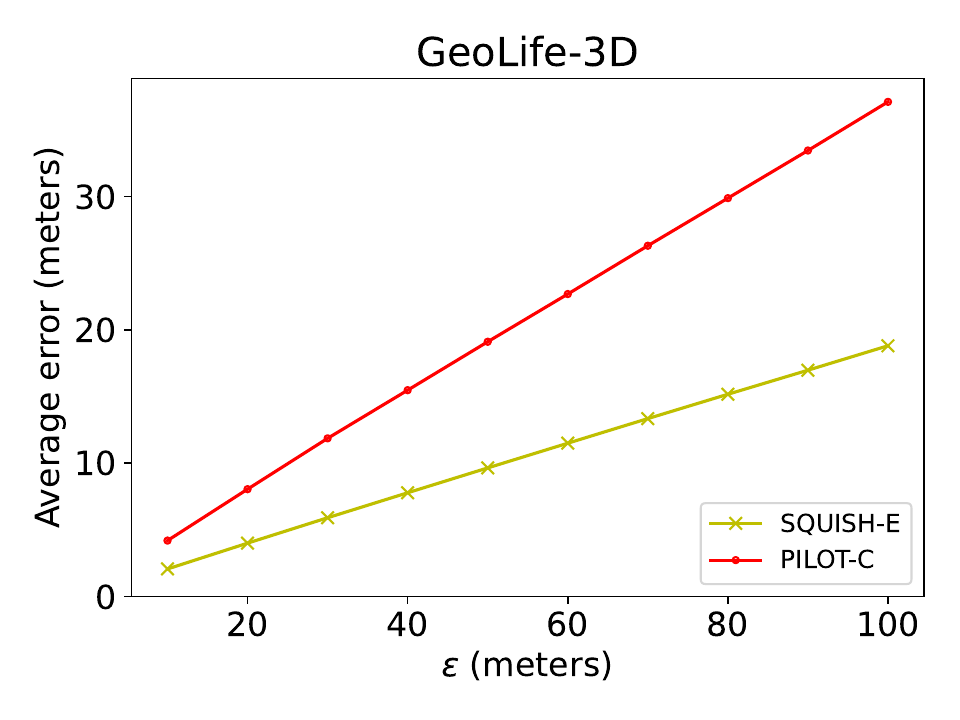}
        \vspace{-0.15in}
        \caption{Average Errors}
    \end{subfigure}
    \vspace{-0.15in}
    \caption{Performance on 3D dataset.
    }
    \vspace{-0.15in}
    \label{fig:Exp-test_error_3d}
\end{figure}

\subsubsection{Performance on 3D trajectories}

In the last set of tests, we evaluate the performance of our algorithm \model\ with SQUISH-E on 3D dataset. 
%
We evaluate how compression performance varies with changes of error bounds, by varying $\epsilon$ from $10$ to $100$ meters on GeoLife-3D. The results are reported in Figure~\ref{fig:Exp-test_error_3d}. On average, \model\ achieves compression ratios of 67.1\% relative to SQUISH-E.

\section{Related Work}
\label{sec:related_work}

In this section, we review prior work related to trajectory compression and the usage of DCT in data compression, both of which are closely connected to this study.

\subsection{Trajectory compression}
Trajectory compression has been extensively studied to minimize storage overhead while ensuring bounded spatio-temporal errors. Due to the limitations of lossless compression algorithms, including low compression ratio and limited efficiency, recent research has focused on lossy compression techniques. These can be broadly categorized into two main classes: geometry-driven methods and semantics-based methods, both of which have significant limitations.

\vspace{-0.08in}
\subsubsection{Geometry-driven approaches} 
Geometry-driven approaches focus on spatial and temporal properties of trajectory data. 
A well-known technique is \textit{trajectory simplification} \cite{cao2003spatio, douglas1973algorithms}, also called line simplification, which selectively discards trajectory points while adhering to specific error metrics such as Perpendicular Euclidean Distance (PED), Synchronized Euclidean Distance (SED), and Direction-Aware Distance (DAD).

A foundational contribution is the ``min-\#'' problem~\cite{lin2021error} introduced by Imai and Iri~\cite{imai1986computational}, which models trajectory simplification as a graph optimization problem, achieving an optimal segmentation strategy with a time complexity of $O(n^3)$. Subsequent improvements reduced this complexity to $O(n^2)$ for PED-based simplification, though scalability remains a challenge for large datasets. 

To address computational constraints, various sub-optimal solutions have been proposed. 
Batch algorithms such as Douglas-Peucker (DP)~\cite{douglas1973algorithms} and Pavlidis's method~\cite{pavlidis1974segmentation}. The former splits the trajectory until it no longer needs to be cut, while the latter merges the trajectory until it no longer needs to be merged. 
Online algorithms do not require any prior knowledge of the entire trajectory. They process trajectory points incrementally with a local buffer. SQUISH-E~\cite{muckell2014compression} uses a priority queue to determine the importance of each point, removing the least significant ones while bounding the overall error.
One-Pass algorithms process each trajectory point once. OPERB~\cite{lin2017one} achieves linear time complexity using PED, whereas CISED~\cite{lin2019one} introduces a spatio-temporal cone intersection technique for SED, achieving near-optimal performance.

Though effective, these trajectory simplification algorithms have common shortcomings since they can hardly balance between compression ratio, average error, efficiency, and the ability to handle multidimensional data. In contrast, our method has achieved near optimal results in all aspects by using frequency-domain, physics-informed compression, treating the trajectory as a continuous curve and optimizing the storage structure.

Other geometry-based techniques include delta compression, where each timestamp in a trajectory is encoded as a delta across time, longitude, and latitude. 
TrajStore~\cite{cudre2010trajstore} framework applies conventional delta compression techniques, incorporating a constant prediction model and leading-zero encoding scheme. Trajic~\cite{nibali2015trajic} refines this approach by introducing a temporally-aware linear predictor, supporting both lossless and lossy compression. Still, these methods generally fall short in compression performance compared to trajectory simplification techniques above and also fail to solve the problem of trajectory continuity.

\subsubsection{Semantics-driven methods} Semantics-driven methods utilize domain-specific knowledge, such as road networks or motion patterns, to enhance trajectory compression quality.
For vehicle trajectories constrained by road networks,  the process typically begins with map-matching~\cite{song2012quick,newson2009hidden}, followed by compression.
For example, COMPRESS~\cite{DBLP:journals/pvldb/SongSZZ14,han2017compress} identifies frequently occurring sub-trajectories and encodes them more efficiently, while UTCQ~\cite{li2020compression} focuses on the compression and queries of uncertain trajectories in road networks. Other methodologies~\cite{richter2012semantic} extend to custom domains by incorporating user-defined domain knowledge to gain additional compression improvement. 

More recent studies have explored learning-based techniques. \cite{wang2021error,wang2021trajectory,fang2023lightweight} RLTS~\cite{wang2021trajectory} formulates online trajectory simplification as a sequential decision process and learns an optimal policy via reinforcement learning.
S3~\cite{fang2023lightweight} builds a context-aware Graph Neural Network (GNN) to capture local and global mobility dependencies and proposes a novel error-distance metric called context-aware error distance (CED). While CED improves semantic fidelity, S3 shows limited gains on conventional error metrics.

Despite their effectiveness in specific domain, these methods often fail to generalize to unconstrained trajectories, such as those from pedestrians or drones, and face difficulties in balancing compression ratio with multidimensional error constraints. Comparing to these methods, our model does not require additional semantic information and can be easily transferred to various datasets.

\subsection{The usage of DCT in data compression}

The Discrete Cosine Transform (DCT) is a foundational technique in modern multimedia compression, renowned for its ability to concentrate signal energy into a small number of low-frequency coefficients. This property enables efficient signal representation and has led to DCT's widespread adoption in image compression (e.g., JPEG-XT~\cite{artusi2019overview} and JPEG~\cite{wallace1992jpeg}), audio compression (e.g., MP3 and AAC~\cite{brandenburg1999mp3}), and video compression (e.g., MPEG \cite{le1991mpeg} and HEVC \cite{sze2014high}). However, these applications typically tolerate globally distributed errors, whereas trajectory compression requires strict local error bounds. This fundamental mismatch has historically limited the application of DCT to spatiotemporal data compression.

Notably absent from the literature is the use of DCT for trajectory compression. This gap stems from two inherent limitations of conventional DCT approaches:
1) Lack of local error control: The global energy compaction characteristic distributes reconstruction errors unevenly, making it difficult to guarantee maximum deviation at individual trajectory points. 
2) Spatiotemporal decoupling: Standard DCT implementations process spatial and temporal dimensions independently, disrupting the intrinsic synchronization between position and timestamp in trajectory data.  

Due to these limitations, trajectory compression has traditionally favored specialized geometric algorithms (e.g., DP variants) and error-bounded temporal methods (e.g., SQUISH-E), rather than frequency-domain approaches like DCT. Our work directly addresses this gap by introducing a novel synchronization-constrained DCT framework that enforces strict per-point error bounds, enabling DCT to be effectively applied to spatiotemporal trajectory data for the first time.

\section{Conclusions}
\label{sec:conclusions}

This paper presents \model, an error-bounded trajectory compression algorithm based on synchronous distance, which is the first model to leverage the physical significance of
motion to guide trajectory compression. Experimental results demonstrate the effectiveness of \model: it achieves an average improvement of 19.2\% in compression ratio compared to CISED-W, the state-of-the-art SED-based line simplification algorithm. Furthermore, \model\ significantly enhances fidelity, reducing the average error by 32.6\% relative to CISED-W. Critically, \model\ extends naturally to three-dimensional trajectories while maintaining its complexity, yielding a substantial 49\% gain in compression ratio over SQUISH-E, the most efficient and effective algorithm for 3D datasets.




\bibliographystyle{ACM-Reference-Format}
\bibliography{reference}


\begin{thebibliography}{39}


\ifx \showCODEN    \undefined \def \showCODEN     #1{\unskip}     \fi
\ifx \showDOI      \undefined \def \showDOI       #1{#1}\fi
\ifx \showISBNx    \undefined \def \showISBNx     #1{\unskip}     \fi
\ifx \showISBNxiii \undefined \def \showISBNxiii  #1{\unskip}     \fi
\ifx \showISSN     \undefined \def \showISSN      #1{\unskip}     \fi
\ifx \showLCCN     \undefined \def \showLCCN      #1{\unskip}     \fi
\ifx \shownote     \undefined \def \shownote      #1{#1}          \fi
\ifx \showarticletitle \undefined \def \showarticletitle #1{#1}   \fi
\ifx \showURL      \undefined \def \showURL       {\relax}        \fi
\providecommand\bibfield[2]{#2}
\providecommand\bibinfo[2]{#2}
\providecommand\natexlab[1]{#1}
\providecommand\showeprint[2][]{arXiv:#2}

\bibitem[Ahmed et~al\mbox{.}(1974)]%
        {DBLP:journals/tc/AhmedNR74}
\bibfield{author}{\bibinfo{person}{Nasir Ahmed}, \bibinfo{person}{T.~Raj Natarajan}, {and} \bibinfo{person}{K.~R. Rao}.} \bibinfo{year}{1974}\natexlab{}.
\newblock \showarticletitle{Discrete Cosine Transform}.
\newblock \bibinfo{journal}{\emph{{IEEE} Trans. Computers}} \bibinfo{volume}{23}, \bibinfo{number}{1} (\bibinfo{year}{1974}), \bibinfo{pages}{90--93}.
\newblock
\urldef\tempurl%
\url{https://doi.org/10.1109/T-C.1974.223784}
\showDOI{\tempurl}


\bibitem[Arai et~al\mbox{.}(1988)]%
        {arai1988fast}
\bibfield{author}{\bibinfo{person}{Yukihiro Arai}, \bibinfo{person}{Takeshi Agui}, {and} \bibinfo{person}{Masayuki Nakajima}.} \bibinfo{year}{1988}\natexlab{}.
\newblock \showarticletitle{A fast DCT-SQ scheme for images}.
\newblock \bibinfo{journal}{\emph{IEICE TRANSACTIONS (1976-1990)}} \bibinfo{volume}{71}, \bibinfo{number}{11} (\bibinfo{year}{1988}), \bibinfo{pages}{1095--1097}.
\newblock


\bibitem[Artusi et~al\mbox{.}(2019)]%
        {artusi2019overview}
\bibfield{author}{\bibinfo{person}{Alessandro Artusi}, \bibinfo{person}{Rafa{\l}~K Mantiuk}, \bibinfo{person}{Thomas Richter}, \bibinfo{person}{Philippe Hanhart}, \bibinfo{person}{Pavel Korshunov}, \bibinfo{person}{Massimiliano Agostinelli}, \bibinfo{person}{Arkady Ten}, {and} \bibinfo{person}{Touradj Ebrahimi}.} \bibinfo{year}{2019}\natexlab{}.
\newblock \showarticletitle{Overview and evaluation of the JPEG XT HDR image compression standard}.
\newblock \bibinfo{journal}{\emph{Journal of Real-Time Image Processing}}  \bibinfo{volume}{16} (\bibinfo{year}{2019}), \bibinfo{pages}{413--428}.
\newblock


\bibitem[Brandenburg(1999)]%
        {brandenburg1999mp3}
\bibfield{author}{\bibinfo{person}{Karlheinz Brandenburg}.} \bibinfo{year}{1999}\natexlab{}.
\newblock \showarticletitle{MP3 and AAC explained}. In \bibinfo{booktitle}{\emph{Audio Engineering Society Conference: 17th International Conference: High-Quality Audio Coding}}. Audio Engineering Society.
\newblock


\bibitem[Caesar et~al\mbox{.}(2021)]%
        {caesar2021nuplan}
\bibfield{author}{\bibinfo{person}{Holger Caesar}, \bibinfo{person}{Juraj Kabzan}, \bibinfo{person}{Kok~Seang Tan}, \bibinfo{person}{Whye~Kit Fong}, \bibinfo{person}{Eric Wolff}, \bibinfo{person}{Alex Lang}, \bibinfo{person}{Luke Fletcher}, \bibinfo{person}{Oscar Beijbom}, {and} \bibinfo{person}{Sammy Omari}.} \bibinfo{year}{2021}\natexlab{}.
\newblock \showarticletitle{nuplan: A closed-loop ml-based planning benchmark for autonomous vehicles}.
\newblock \bibinfo{journal}{\emph{arXiv preprint arXiv:2106.11810}} (\bibinfo{year}{2021}).
\newblock


\bibitem[Cao et~al\mbox{.}(2003)]%
        {cao2003spatio}
\bibfield{author}{\bibinfo{person}{Hu Cao}, \bibinfo{person}{Ouri Wolfson}, {and} \bibinfo{person}{Goce Trajcevski}.} \bibinfo{year}{2003}\natexlab{}.
\newblock \showarticletitle{Spatio-temporal data reduction with deterministic error bounds}. In \bibinfo{booktitle}{\emph{Proceedings of the 2003 joint workshop on Foundations of mobile computing}}. \bibinfo{pages}{33--42}.
\newblock


\bibitem[Chen et~al\mbox{.}(2012)]%
        {chen2012fast}
\bibfield{author}{\bibinfo{person}{Minjie Chen}, \bibinfo{person}{Mantao Xu}, {and} \bibinfo{person}{Pasi Franti}.} \bibinfo{year}{2012}\natexlab{}.
\newblock \showarticletitle{A fast $ o (n) $ multiresolution polygonal approximation algorithm for GPS trajectory simplification}.
\newblock \bibinfo{journal}{\emph{IEEE Transactions on Image Processing}} \bibinfo{volume}{21}, \bibinfo{number}{5} (\bibinfo{year}{2012}), \bibinfo{pages}{2770--2785}.
\newblock


\bibitem[Chen et~al\mbox{.}(2009)]%
        {chen2009trajectory}
\bibfield{author}{\bibinfo{person}{Yukun Chen}, \bibinfo{person}{Kai Jiang}, \bibinfo{person}{Yu Zheng}, \bibinfo{person}{Chunping Li}, {and} \bibinfo{person}{Nenghai Yu}.} \bibinfo{year}{2009}\natexlab{}.
\newblock \showarticletitle{Trajectory simplification method for location-based social networking services}. In \bibinfo{booktitle}{\emph{Proceedings of the 2009 international workshop on location based social networks}}. \bibinfo{pages}{33--40}.
\newblock


\bibitem[Cudre-Mauroux et~al\mbox{.}(2010)]%
        {cudre2010trajstore}
\bibfield{author}{\bibinfo{person}{Philippe Cudre-Mauroux}, \bibinfo{person}{Eugene Wu}, {and} \bibinfo{person}{Samuel Madden}.} \bibinfo{year}{2010}\natexlab{}.
\newblock \showarticletitle{Trajstore: An adaptive storage system for very large trajectory data sets}. In \bibinfo{booktitle}{\emph{2010 IEEE 26th International Conference on Data Engineering (ICDE 2010)}}. IEEE, \bibinfo{pages}{109--120}.
\newblock


\bibitem[Douglas and Peucker(1973)]%
        {douglas1973algorithms}
\bibfield{author}{\bibinfo{person}{David~H Douglas} {and} \bibinfo{person}{Thomas~K Peucker}.} \bibinfo{year}{1973}\natexlab{}.
\newblock \showarticletitle{Algorithms for the reduction of the number of points required to represent a digitized line or its caricature}.
\newblock \bibinfo{journal}{\emph{Cartographica: the international journal for geographic information and geovisualization}} \bibinfo{volume}{10}, \bibinfo{number}{2} (\bibinfo{year}{1973}), \bibinfo{pages}{112--122}.
\newblock


\bibitem[Fang et~al\mbox{.}(2023)]%
        {fang2023lightweight}
\bibfield{author}{\bibinfo{person}{Ziquan Fang}, \bibinfo{person}{Changhao He}, \bibinfo{person}{Lu Chen}, \bibinfo{person}{Danlei Hu}, \bibinfo{person}{Qichen Sun}, \bibinfo{person}{Linsen Li}, {and} \bibinfo{person}{Yunjun Gao}.} \bibinfo{year}{2023}\natexlab{}.
\newblock \showarticletitle{A lightweight framework for fast trajectory simplification}. In \bibinfo{booktitle}{\emph{2023 IEEE 39th International Conference on Data Engineering (ICDE)}}. IEEE, \bibinfo{pages}{2386--2399}.
\newblock


\bibitem[Gotsman and Kanza(2015)]%
        {gotsman2015dilution}
\bibfield{author}{\bibinfo{person}{Ranit Gotsman} {and} \bibinfo{person}{Yaron Kanza}.} \bibinfo{year}{2015}\natexlab{}.
\newblock \showarticletitle{A dilution-matching-encoding compaction of trajectories over road networks}.
\newblock \bibinfo{journal}{\emph{GeoInformatica}}  \bibinfo{volume}{19} (\bibinfo{year}{2015}), \bibinfo{pages}{331--364}.
\newblock


\bibitem[Han et~al\mbox{.}(2017)]%
        {han2017compress}
\bibfield{author}{\bibinfo{person}{Yunheng Han}, \bibinfo{person}{Weiwei Sun}, {and} \bibinfo{person}{Baihua Zheng}.} \bibinfo{year}{2017}\natexlab{}.
\newblock \showarticletitle{COMPRESS: A comprehensive framework of trajectory compression in road networks}.
\newblock \bibinfo{journal}{\emph{ACM Transactions on Database Systems (TODS)}} \bibinfo{volume}{42}, \bibinfo{number}{2} (\bibinfo{year}{2017}), \bibinfo{pages}{1--49}.
\newblock


\bibitem[Imai and Iri(1986)]%
        {imai1986computational}
\bibfield{author}{\bibinfo{person}{Hiroshi Imai} {and} \bibinfo{person}{Masao Iri}.} \bibinfo{year}{1986}\natexlab{}.
\newblock \showarticletitle{Computational-geometric methods for polygonal approximations of a curve}.
\newblock \bibinfo{journal}{\emph{Computer Vision, Graphics, and Image Processing}} \bibinfo{volume}{36}, \bibinfo{number}{1} (\bibinfo{year}{1986}), \bibinfo{pages}{31--41}.
\newblock


\bibitem[Le~Gall(1991)]%
        {le1991mpeg}
\bibfield{author}{\bibinfo{person}{Didier Le~Gall}.} \bibinfo{year}{1991}\natexlab{}.
\newblock \showarticletitle{MPEG: A video compression standard for multimedia applications}.
\newblock \bibinfo{journal}{\emph{Commun. ACM}} \bibinfo{volume}{34}, \bibinfo{number}{4} (\bibinfo{year}{1991}), \bibinfo{pages}{46--58}.
\newblock


\bibitem[Li et~al\mbox{.}(2020)]%
        {li2020compression}
\bibfield{author}{\bibinfo{person}{Tianyi Li}, \bibinfo{person}{Ruikai Huang}, \bibinfo{person}{Lu Chen}, \bibinfo{person}{Christian~S Jensen}, {and} \bibinfo{person}{Torben~Bach Pedersen}.} \bibinfo{year}{2020}\natexlab{}.
\newblock \showarticletitle{Compression of uncertain trajectories in road networks}.
\newblock \bibinfo{journal}{\emph{Proceedings of the VLDB Endowment}} \bibinfo{volume}{13}, \bibinfo{number}{7} (\bibinfo{year}{2020}), \bibinfo{pages}{1050--1063}.
\newblock


\bibitem[Lin et~al\mbox{.}(2019)]%
        {lin2019one}
\bibfield{author}{\bibinfo{person}{Xuelian Lin}, \bibinfo{person}{Jiahao Jiang}, \bibinfo{person}{Shuai Ma}, \bibinfo{person}{Yimeng Zuo}, {and} \bibinfo{person}{Chunming Hu}.} \bibinfo{year}{2019}\natexlab{}.
\newblock \showarticletitle{One-pass trajectory simplification using the synchronous Euclidean distance}.
\newblock \bibinfo{journal}{\emph{The VLDB Journal}} \bibinfo{volume}{28}, \bibinfo{number}{6} (\bibinfo{year}{2019}), \bibinfo{pages}{897--921}.
\newblock


\bibitem[Lin et~al\mbox{.}(2021)]%
        {lin2021error}
\bibfield{author}{\bibinfo{person}{Xuelian Lin}, \bibinfo{person}{Shuai Ma}, \bibinfo{person}{Jiahao Jiang}, \bibinfo{person}{Yanchen Hou}, {and} \bibinfo{person}{Tianyu Wo}.} \bibinfo{year}{2021}\natexlab{}.
\newblock \showarticletitle{Error bounded line simplification algorithms for trajectory compression: An experimental evaluation}.
\newblock \bibinfo{journal}{\emph{ACM Transactions on Database Systems (TODS)}} \bibinfo{volume}{46}, \bibinfo{number}{3} (\bibinfo{year}{2021}), \bibinfo{pages}{1--44}.
\newblock


\bibitem[Lin et~al\mbox{.}(2017)]%
        {lin2017one}
\bibfield{author}{\bibinfo{person}{Xuelian Lin}, \bibinfo{person}{Shuai Ma}, \bibinfo{person}{Han Zhang}, \bibinfo{person}{Tianyu Wo}, {and} \bibinfo{person}{Jinpeng Huai}.} \bibinfo{year}{2017}\natexlab{}.
\newblock \showarticletitle{One-pass error bounded trajectory simplification}.
\newblock \bibinfo{journal}{\emph{arXiv preprint arXiv:1702.05597}} (\bibinfo{year}{2017}).
\newblock


\bibitem[Liu et~al\mbox{.}(2015)]%
        {liu2015bounded}
\bibfield{author}{\bibinfo{person}{Jiajun Liu}, \bibinfo{person}{Kun Zhao}, \bibinfo{person}{Philipp Sommer}, \bibinfo{person}{Shuo Shang}, \bibinfo{person}{Brano Kusy}, {and} \bibinfo{person}{Raja Jurdak}.} \bibinfo{year}{2015}\natexlab{}.
\newblock \showarticletitle{Bounded quadrant system: Error-bounded trajectory compression on the go}. In \bibinfo{booktitle}{\emph{2015 IEEE 31st International Conference on Data Engineering}}. IEEE, \bibinfo{pages}{987--998}.
\newblock


\bibitem[Liu et~al\mbox{.}(2016)]%
        {liu2016novel}
\bibfield{author}{\bibinfo{person}{Jiajun Liu}, \bibinfo{person}{Kun Zhao}, \bibinfo{person}{Philipp Sommer}, \bibinfo{person}{Shuo Shang}, \bibinfo{person}{Brano Kusy}, \bibinfo{person}{Jae-Gil Lee}, {and} \bibinfo{person}{Raja Jurdak}.} \bibinfo{year}{2016}\natexlab{}.
\newblock \showarticletitle{A novel framework for online amnesic trajectory compression in resource-constrained environments}.
\newblock \bibinfo{journal}{\emph{IEEE Transactions on knowledge and data engineering}} \bibinfo{volume}{28}, \bibinfo{number}{11} (\bibinfo{year}{2016}), \bibinfo{pages}{2827--2841}.
\newblock


\bibitem[Loeffler et~al\mbox{.}(1989)]%
        {loeffler1989practical}
\bibfield{author}{\bibinfo{person}{Christoph Loeffler}, \bibinfo{person}{Adriaan Ligtenberg}, {and} \bibinfo{person}{George~S Moschytz}.} \bibinfo{year}{1989}\natexlab{}.
\newblock \showarticletitle{Practical fast 1-D DCT algorithms with 11 multiplications}. In \bibinfo{booktitle}{\emph{International Conference on Acoustics, Speech, and Signal Processing,}}. IEEE, \bibinfo{pages}{988--991}.
\newblock


\bibitem[Mariescu-Istodor and Fr{\"a}nti(2017)]%
        {mariescu2017grid}
\bibfield{author}{\bibinfo{person}{Radu Mariescu-Istodor} {and} \bibinfo{person}{Pasi Fr{\"a}nti}.} \bibinfo{year}{2017}\natexlab{}.
\newblock \showarticletitle{Grid-based method for GPS route analysis for retrieval}.
\newblock \bibinfo{journal}{\emph{ACM Transactions on Spatial Algorithms and Systems (TSAS)}} \bibinfo{volume}{3}, \bibinfo{number}{3} (\bibinfo{year}{2017}), \bibinfo{pages}{1--28}.
\newblock


\bibitem[Meratnia and de~By(2004)]%
        {meratnia2004spatiotemporal}
\bibfield{author}{\bibinfo{person}{Nirvana Meratnia} {and} \bibinfo{person}{Rolf~A de By}.} \bibinfo{year}{2004}\natexlab{}.
\newblock \showarticletitle{Spatiotemporal compression techniques for moving point objects}. In \bibinfo{booktitle}{\emph{International Conference on Extending Database Technology}}. Springer, \bibinfo{pages}{765--782}.
\newblock


\bibitem[Muckell et~al\mbox{.}(2014)]%
        {muckell2014compression}
\bibfield{author}{\bibinfo{person}{Jonathan Muckell}, \bibinfo{person}{Paul~W Olsen}, \bibinfo{person}{Jeong-Hyon Hwang}, \bibinfo{person}{Catherine~T Lawson}, {and} \bibinfo{person}{SS Ravi}.} \bibinfo{year}{2014}\natexlab{}.
\newblock \showarticletitle{Compression of trajectory data: a comprehensive evaluation and new approach}.
\newblock \bibinfo{journal}{\emph{GeoInformatica}}  \bibinfo{volume}{18} (\bibinfo{year}{2014}), \bibinfo{pages}{435--460}.
\newblock


\bibitem[Newson and Krumm(2009)]%
        {newson2009hidden}
\bibfield{author}{\bibinfo{person}{Paul Newson} {and} \bibinfo{person}{John Krumm}.} \bibinfo{year}{2009}\natexlab{}.
\newblock \showarticletitle{Hidden Markov map matching through noise and sparseness}. In \bibinfo{booktitle}{\emph{Proceedings of the 17th ACM SIGSPATIAL international conference on advances in geographic information systems}}. \bibinfo{pages}{336--343}.
\newblock


\bibitem[Nibali and He(2015)]%
        {nibali2015trajic}
\bibfield{author}{\bibinfo{person}{Aiden Nibali} {and} \bibinfo{person}{Zhen He}.} \bibinfo{year}{2015}\natexlab{}.
\newblock \showarticletitle{Trajic: An effective compression system for trajectory data}.
\newblock \bibinfo{journal}{\emph{IEEE Transactions on Knowledge and Data Engineering}} \bibinfo{volume}{27}, \bibinfo{number}{11} (\bibinfo{year}{2015}), \bibinfo{pages}{3138--3151}.
\newblock


\bibitem[Pavlidis and Horowitz(1974)]%
        {pavlidis1974segmentation}
\bibfield{author}{\bibinfo{person}{Theodosios Pavlidis} {and} \bibinfo{person}{Steven~L Horowitz}.} \bibinfo{year}{1974}\natexlab{}.
\newblock \showarticletitle{Segmentation of plane curves}.
\newblock \bibinfo{journal}{\emph{IEEE transactions on Computers}} \bibinfo{volume}{100}, \bibinfo{number}{8} (\bibinfo{year}{1974}), \bibinfo{pages}{860--870}.
\newblock


\bibitem[Richter et~al\mbox{.}(2012)]%
        {richter2012semantic}
\bibfield{author}{\bibinfo{person}{Kai-Florian Richter}, \bibinfo{person}{Falko Schmid}, {and} \bibinfo{person}{Patrick Laube}.} \bibinfo{year}{2012}\natexlab{}.
\newblock \showarticletitle{Semantic trajectory compression: Representing urban movement in a nutshell}.
\newblock \bibinfo{journal}{\emph{Journal of Spatial Information Science}} \bibinfo{number}{4} (\bibinfo{year}{2012}), \bibinfo{pages}{3--30}.
\newblock


\bibitem[Sandu~Popa et~al\mbox{.}(2015)]%
        {sandu2015spatio}
\bibfield{author}{\bibinfo{person}{Iulian Sandu~Popa}, \bibinfo{person}{Karine Zeitouni}, \bibinfo{person}{Vincent Oria}, {and} \bibinfo{person}{Ahmed Kharrat}.} \bibinfo{year}{2015}\natexlab{}.
\newblock \showarticletitle{Spatio-temporal compression of trajectories in road networks}.
\newblock \bibinfo{journal}{\emph{GeoInformatica}}  \bibinfo{volume}{19} (\bibinfo{year}{2015}), \bibinfo{pages}{117--145}.
\newblock


\bibitem[Song et~al\mbox{.}(2012)]%
        {song2012quick}
\bibfield{author}{\bibinfo{person}{Renchu Song}, \bibinfo{person}{Wei Lu}, \bibinfo{person}{Weiwei Sun}, \bibinfo{person}{Yan Huang}, {and} \bibinfo{person}{Chunan Chen}.} \bibinfo{year}{2012}\natexlab{}.
\newblock \showarticletitle{Quick map matching using multi-core cpus}. In \bibinfo{booktitle}{\emph{Proceedings of the 20th International Conference on Advances in Geographic Information Systems}}. \bibinfo{pages}{605--608}.
\newblock


\bibitem[Song et~al\mbox{.}(2014)]%
        {DBLP:journals/pvldb/SongSZZ14}
\bibfield{author}{\bibinfo{person}{Renchu Song}, \bibinfo{person}{Weiwei Sun}, \bibinfo{person}{Baihua Zheng}, {and} \bibinfo{person}{Yu Zheng}.} \bibinfo{year}{2014}\natexlab{}.
\newblock \showarticletitle{{PRESS:} {A} Novel Framework of Trajectory Compression in Road Networks}.
\newblock \bibinfo{journal}{\emph{Proc. {VLDB} Endow.}} \bibinfo{volume}{7}, \bibinfo{number}{9} (\bibinfo{year}{2014}), \bibinfo{pages}{661--672}.
\newblock
\urldef\tempurl%
\url{https://doi.org/10.14778/2732939.2732940}
\showDOI{\tempurl}


\bibitem[Sze et~al\mbox{.}(2014)]%
        {sze2014high}
\bibfield{author}{\bibinfo{person}{Vivienne Sze}, \bibinfo{person}{Madhukar Budagavi}, {and} \bibinfo{person}{Gary~J Sullivan}.} \bibinfo{year}{2014}\natexlab{}.
\newblock \showarticletitle{High efficiency video coding (HEVC)}.
\newblock \bibinfo{journal}{\emph{Integrated circuit and systems, algorithms and architectures}}  \bibinfo{volume}{39} (\bibinfo{year}{2014}), \bibinfo{pages}{40}.
\newblock


\bibitem[Wallace(1992)]%
        {wallace1992jpeg}
\bibfield{author}{\bibinfo{person}{Gregory~K Wallace}.} \bibinfo{year}{1992}\natexlab{}.
\newblock \showarticletitle{The JPEG still picture compression standard}.
\newblock \bibinfo{journal}{\emph{IEEE transactions on consumer electronics}} \bibinfo{volume}{38}, \bibinfo{number}{1} (\bibinfo{year}{1992}), \bibinfo{pages}{xviii--xxxiv}.
\newblock


\bibitem[Wang et~al\mbox{.}(2021a)]%
        {wang2021trajectory}
\bibfield{author}{\bibinfo{person}{Zheng Wang}, \bibinfo{person}{Cheng Long}, {and} \bibinfo{person}{Gao Cong}.} \bibinfo{year}{2021}\natexlab{a}.
\newblock \showarticletitle{Trajectory simplification with reinforcement learning}. In \bibinfo{booktitle}{\emph{2021 IEEE 37th International Conference on Data Engineering (ICDE)}}. IEEE, \bibinfo{pages}{684--695}.
\newblock


\bibitem[Wang et~al\mbox{.}(2021b)]%
        {wang2021error}
\bibfield{author}{\bibinfo{person}{Zheng Wang}, \bibinfo{person}{Cheng Long}, \bibinfo{person}{Gao Cong}, {and} \bibinfo{person}{Qianru Zhang}.} \bibinfo{year}{2021}\natexlab{b}.
\newblock \showarticletitle{Error-bounded online trajectory simplification with multi-agent reinforcement learning}. In \bibinfo{booktitle}{\emph{Proceedings of the 27th ACM SIGKDD conference on knowledge discovery \& data mining}}. \bibinfo{pages}{1758--1768}.
\newblock


\bibitem[Yang et~al\mbox{.}(2018)]%
        {DBLP:journals/tkde/YangWYLZ18}
\bibfield{author}{\bibinfo{person}{Xiaochun Yang}, \bibinfo{person}{Bin Wang}, \bibinfo{person}{Kai Yang}, \bibinfo{person}{Chengfei Liu}, {and} \bibinfo{person}{Baihua Zheng}.} \bibinfo{year}{2018}\natexlab{}.
\newblock \showarticletitle{A Novel Representation and Compression for Queries on Trajectories in Road Networks}.
\newblock \bibinfo{journal}{\emph{{IEEE} Trans. Knowl. Data Eng.}} \bibinfo{volume}{30}, \bibinfo{number}{4} (\bibinfo{year}{2018}), \bibinfo{pages}{613--629}.
\newblock
\urldef\tempurl%
\url{https://doi.org/10.1109/TKDE.2017.2776927}
\showDOI{\tempurl}


\bibitem[Zhao et~al\mbox{.}(2018)]%
        {zhao2018rest}
\bibfield{author}{\bibinfo{person}{Yan Zhao}, \bibinfo{person}{Shuo Shang}, \bibinfo{person}{Yu Wang}, \bibinfo{person}{Bolong Zheng}, \bibinfo{person}{Quoc Viet~Hung Nguyen}, {and} \bibinfo{person}{Kai Zheng}.} \bibinfo{year}{2018}\natexlab{}.
\newblock \showarticletitle{Rest: A reference-based framework for spatio-temporal trajectory compression}. In \bibinfo{booktitle}{\emph{Proceedings of the 24th ACM SIGKDD international conference on knowledge discovery \& data mining}}. \bibinfo{pages}{2797--2806}.
\newblock


\bibitem[Zheng et~al\mbox{.}(2010)]%
        {zheng2010geolife}
\bibfield{author}{\bibinfo{person}{Yu Zheng}, \bibinfo{person}{Xing Xie}, \bibinfo{person}{Wei-Ying Ma}, {et~al\mbox{.}}} \bibinfo{year}{2010}\natexlab{}.
\newblock \showarticletitle{GeoLife: A collaborative social networking service among user, location and trajectory.}
\newblock \bibinfo{journal}{\emph{IEEE Data Eng. Bull.}} \bibinfo{volume}{33}, \bibinfo{number}{2} (\bibinfo{year}{2010}), \bibinfo{pages}{32--39}.
\newblock


\end{thebibliography}

\clearpage

\end{document}